\documentclass[lettersize,journal]{IEEEtran}
\usepackage{amsmath,amsfonts}
\usepackage{algorithmic}
\usepackage{algorithm}
\usepackage{array}
\usepackage{multirow}
\usepackage[caption=false,font=normalsize,labelfont=sf,textfont=sf]{subfig}
\usepackage{textcomp}
\usepackage{stfloats}
\usepackage{url}
\usepackage{verbatim}
\usepackage{graphicx}
\usepackage{cite}
\begin{document}

\title{MBTI: A Multi-Branch Efficient Fine-Tuning Framework for Hyperspectral Image Classification with Foundation Models}

\author{Mingzhen Xu, Haonan Guo, Di Wang, Yinghua Qu, Zhiliang Zhou, Lei Zhang, Huiwen Yao, Rui Zhao, Fengxiang Wang, Gang Wan, Bo Du, and Liangpei Zhang,~\IEEEmembership{Fellow,~IEEE}

\thanks{
Corresponding authors: Di Wang, Fengxiang Wang, Gang Wan, and Bo Du
}
\thanks{Mingzhen Xu, Di Wang, and Bo Du are with the School of Computer Science, Wuhan University, Wuhan 430072, China. Di Wang, and Bo Du are also with the Zhongguancun Academy, Beijing 100094, China (e-mail: mingzhenxu@whu.edu.cn; d\_wang@whu.edu.cn; dubo@whu.edu.cn)}
\thanks{Haonan Guo and Liangpei Zhang are with the State Key Laboratory of Information Engineering in Surveying, Mapping and Remote Sensing, Wuhan University, Wuhan 430079, China. (e-mail: haonan.guo@whu.edu.cn; zlp62@whu.edu.cn)}
\thanks{Yinghua Qu and Huiwen Yao are with the China Petroleum Pipeline Engineering Corporation, Langfang 065000, China. And Huiwen Yao is also with the Hebei Key Laboratory of Underground Energy Storage Technology, Langfang 065000, China. (e-mail: quyinghua@cppe.com.cn; yaohuiwen@cppe.com.cn)}
\thanks{Zhiliang Zhou and Lei Zhang are with the No. 7 Oil Production Plant, Changqing Oilfield Branch, PetroChina Company Limited, Xi'an 710018, China. (e-mail: zhoulz\_cq@petrochina.com.cn; zl17\_cq@petrochina.com.cn)}
\thanks{Rui Zhao is with the Faculty of Electrical Engineering and Computer Science, Ningbo University, Ningbo 315211, China. (e-mail: zhaorui@nbu.edu.cn)}
\thanks{Fengxiang Wang is with the College of Computer Science and Technology, National University of Defense Technology, Changsha 410000, China. (e-mail: wfx23@nudt.edu.cn)}
\thanks{Gang Wan is with the School of Aerospace Information, Space Engineering University, Beijing 101400, China, and also with the Key Laboratory of Intelligent Processing and Application Technology of Satellite Information, Beijing 101400, China (e-mail: wangang@hgd.edu.cn)}
}

\markboth{IEEE TRANSACTIONS ON GEOSCIENCE AND REMOTE SENSING}%
{Xu \MakeLowercase{\textit{et al.}}: MBTI for Hyperspectral Image Classification}

\maketitle

\begin{abstract}
Hyperspectral foundation models learn transferable spectral-spatial representations from large-scale unlabeled data. They provide an effective paradigm for adapting to downstream hyperspectral image (HSI) classification tasks with limited labeled samples. However, spectral band configurations vary substantially across sensors, which makes direct model transfer difficult. Existing adaptation strategies often compress, select, or reshape the original spectra to match model-specific input requirements. These operations may discard useful spectral information and weaken local spectral continuity. To address this problem, we propose MBTI, a multi-branch efficient fine-tuning framework for hyperspectral image classification. MBTI adapts hyperspectral foundation models to downstream classification tasks while preserving full-band spectral information. First, we introduce a spectral-continuity-preserving multi-branch preprocessing strategy. The original HSI is divided into multiple continuous spectral subsets, and a band reuse mechanism is used when the remaining bands cannot form a complete branch. This avoids invalid padding and unnecessary spectral loss. Second, independent Low-Rank Adaptation (LoRA) modules are inserted into each branch. They enable different spectral intervals to learn task-specific discriminative features while keeping most pre-trained parameters frozen. Finally, a multi-branch channel attention fusion module adaptively recalibrates and integrates features from all spectral branches. Experiments on three public hyperspectral datasets show that MBTI achieves competitive and superior performance compared with representative classification methods. Under the final rank-8 configuration, only about 2.33\%--2.36\% of the parameters are trainable. The source code is available at https://github.com/Azhenmiddleblock/MBTI.
\end{abstract}

\begin{IEEEkeywords}
Hyperspectral image classification, hyperspectral foundation models, parameter-efficient fine-tuning, remote sensing.
\end{IEEEkeywords}

\section{Introduction}
\IEEEPARstart{H}{yperspectral} images (HSIs) are usually represented as three-dimensional data cubes. They contain rich spatial structures and continuous spectral responses across dozens to hundreds of narrow bands. Compared with conventional RGB imagery, HSIs provide dense and contiguous spectral measurements. These measurements capture detailed material-specific signatures and make HSIs particularly suitable for accurate land-cover classification of spectrally similar materials \cite{ref1,ref2,ref3}.

HSI classification aims to assign a semantic label to each pixel according to its spectral-spatial characteristics. Early studies mainly relied on conventional machine learning methods, such as Support Vector Machine (SVM) \cite{ref4} and Random Forest (RF) \cite{ref5}. These methods can be effective when training samples are limited. However, their performance is often constrained by handcrafted features, shallow representations, and limited capability in modeling complex spectral-spatial dependencies. The high dimensionality and strong band correlation of HSIs further make it difficult for these methods to fully exploit discriminative information from both domains.

With the development of deep learning, LSTM \cite{ref6}, CNN \cite{ref7}, and Transformer-based models \cite{ref8} have been widely introduced into HSI classification. These models learn spectral-spatial representations and have achieved substantial improvements on benchmark datasets. However, their strong representation capability usually depends on sufficient labeled samples. In practical HSI applications, high-quality annotations are expensive and time-consuming to obtain. Training samples are therefore persistently scarce \cite{ref9}. This limitation has motivated studies on cross-domain few-shot learning (FSL), where knowledge learned from label-rich source domains is transferred to label-scarce target domains \cite{ref10}.

FSL has gained widespread attention because it can recognize classes with only a few labeled samples \cite{ref11}. It is often formulated as a meta-learning paradigm. The goal is to learn transferable knowledge that generalizes from source tasks to target tasks. This is typically achieved by constructing episodic training tasks across source and target domains, enabling the model to learn discriminative embeddings and domain-invariant meta-knowledge \cite{ref12}. Based on this paradigm, Wang et al. \cite{ref13} designed a local spectrum-spatial alignment module to balance domain-invariant and domain-discriminative representations. Qin et al. \cite{ref14} proposed a prototype rectification network to alleviate biased prototypes and domain shifts. Liu et al. \cite{ref15} further proposed a category-specific prototype self-refinement module, which adaptively updates class prototypes using information from the query set. Although these FSL methods improve classification performance in label-scarce scenarios, they still rely on carefully designed episodic training, source-target task construction, and prototype adaptation. Their performance may be affected by domain shifts, category discrepancies, and the limited generality of task-specific transfer models.

In recent years, foundation models \cite{ref16} have provided a new paradigm for visual representation learning. Unlike conventional task-specific models, they are pre-trained on large-scale data and can be adapted to downstream tasks with limited labeled samples through fine-tuning or prompting \cite{ref17}. In hyperspectral remote sensing, HyperSIGMA \cite{ref18} was developed for hyperspectral intelligence comprehension and demonstrated strong performance across multiple HSI interpretation tasks. Following this line of research, several studies have further explored how hyperspectral foundation models can handle heterogeneous spectral configurations. HyperSL \cite{ref19} introduced a spectral foundation model that treats hyperspectral signals as sequential tokens and incorporates wavelength information into positional encoding, enabling knowledge transfer across datasets with different spectral ranges and band numbers. HySens \cite{ref20} developed a sensor-agnostic hyperspectral foundation model by combining a spectral harmonizer with wavelength-based spectral embeddings, improving the adaptability of pretrained models to heterogeneous hyperspectral sensors. From the perspective of downstream adaptation, HyperFree \cite{ref21} adopted a channel-adaptive and tuning-free strategy to reduce the cost of applying hyperspectral foundation models to downstream tasks. These studies demonstrate the potential of foundation models for HSI analysis. However, their adaptation to classification datasets with diverse spectral band configurations remains insufficiently explored.

HSIs acquired by different sensors often differ in band number, wavelength coverage, spectral response functions, and band intervals. When a pre-trained model has specific input requirements, downstream data are often compressed, selected, interpolated, or reshaped before being fed into the model. Such preprocessing may discard useful spectral information or weaken the local continuity of the original spectra. This is particularly undesirable for HSI classification, where subtle spectral differences are often critical.

\begin{figure*}[!t]
\centering
\includegraphics[width=\textwidth]{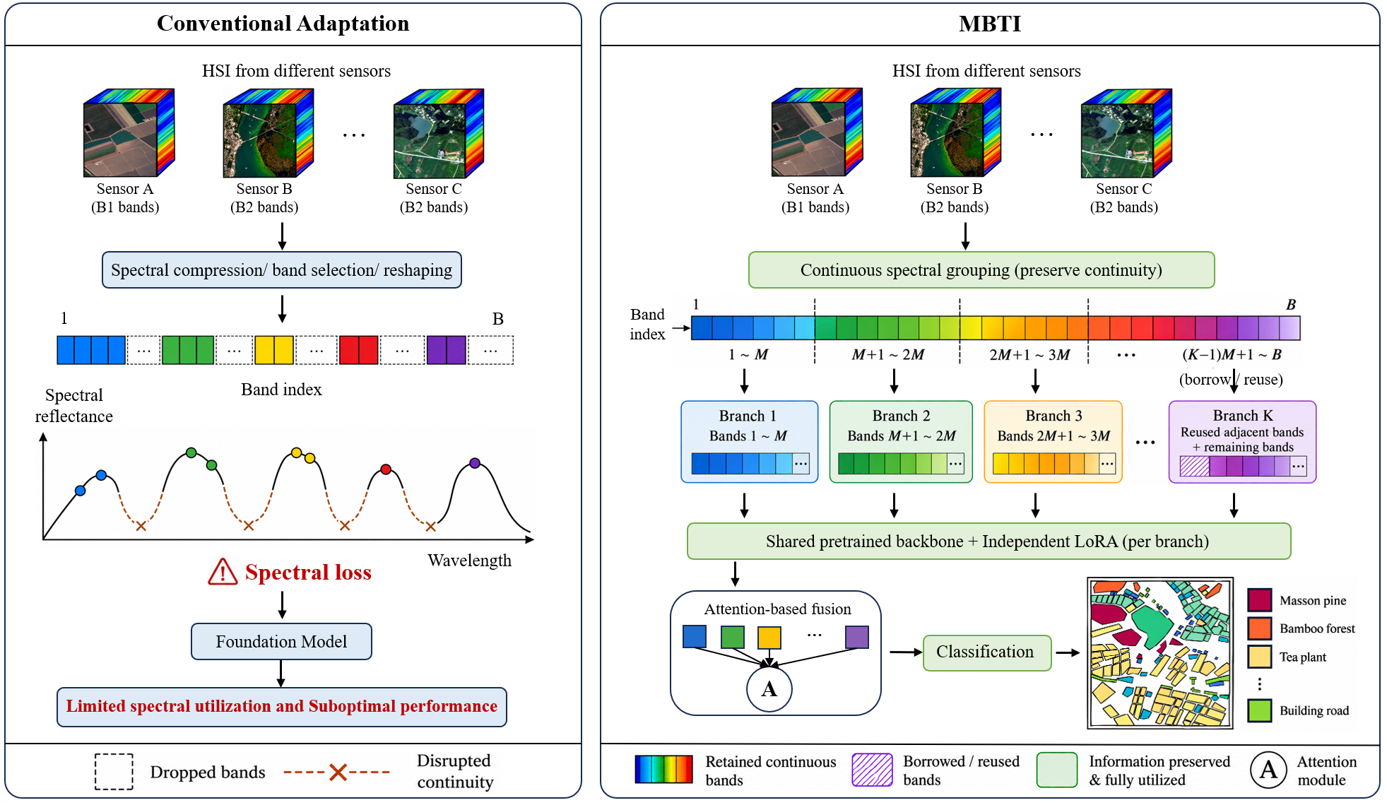}
\caption{Spectral information loss in conventional adaptation vs. spectral-continuity-preserving multi-branch adaptation.}
\label{fig:mbti_overview}
\end{figure*}

To address this issue, we propose a multi-branch efficient fine-tuning framework for hyperspectral image classification, termed MBTI. Its core idea is to preserve the original spectral information while adapting HSIs with different band configurations to a pre-trained foundation model. Specifically, the full-band HSI is divided into several continuous spectral groups. When the remaining bands are insufficient to form a complete group, a band reuse strategy constructs a valid continuous input branch without zero padding or band discarding. As shown in Fig.~\ref{fig:mbti_overview}, each spectral group is then fed into an independent branch built upon the pre-trained foundation model.

Although the multi-branch design preserves spectral integrity, directly fine-tuning all branches would introduce substantial computational and optimization burdens. We therefore introduce parameter-efficient fine-tuning into MBTI. The pre-trained backbone is frozen, and independent Low-Rank Adaptation (LoRA) modules are inserted into each branch. LoRA approximates the update of a large weight matrix with two trainable low-rank matrices, denoted as $A$ and $B$. This design greatly reduces the number of trainable parameters while retaining the adaptation capability of each branch.

Finally, features extracted from different spectral branches must be integrated for classification. Different spectral intervals contribute unequally to different land-cover categories. We therefore design a multi-branch channel attention fusion module to recalibrate and integrate branch features adaptively. This module enhances informative spectral responses and suppresses redundant or less relevant ones.

The main contributions of this paper are summarized as follows:
\begin{enumerate}
\item To alleviate the spectral information loss caused by conventional input adaptation strategies, we propose MBTI, a multi-branch efficient fine-tuning framework based on hyperspectral foundation models. It adapts HSIs with different band configurations while preserving full-band spectral information and local spectral continuity.
\item We design a spectral grouping LoRA strategy. Continuous spectral subsets are processed by independent foundation-model branches equipped with branch-specific LoRA modules. This allows different spectral intervals to learn discriminative task-specific features with only a small number of trainable parameters.
\item We introduce a multi-branch channel attention fusion module to adaptively integrate features from different spectral branches. Experiments on three representative HSI datasets demonstrate the effectiveness and efficiency of the proposed MBTI framework.
\end{enumerate}

The remainder of this paper is organized as follows. Section~II reviews related work. Section~III presents the proposed MBTI framework. Section~IV reports the experimental results and analysis. Section~V concludes this paper.

\section{Related Work}
\subsection{Hyperspectral Foundation Models}
The success of the pre-training and fine-tuning paradigm in natural language processing and computer vision has motivated the development of foundation models in remote sensing \cite{ref22}. By learning transferable representations from large-scale unlabeled or weakly labeled data, foundation models provide a promising solution for downstream tasks where labeled samples are limited. Compared with conventional task-specific models, foundation models are expected to offer stronger generalization and better adaptability across diverse scenes and sensors.

In the hyperspectral community, recent studies have begun to explore foundation models that can exploit rich spectral-spatial information. HyperSIGMA \cite{ref18} was developed as a hyperspectral intelligence comprehension foundation model and demonstrated strong representation capability across multiple HSI interpretation tasks. SpectralEarth \cite{ref23} constructed a large-scale EnMAP-based hyperspectral dataset and pretrained foundation models with self-supervised learning, providing an important benchmark for large-scale HSI representation learning. HyperSL \cite{ref19} regarded hyperspectral signals as sequential spectral tokens and introduced wavelength-based positional encoding to align spectral features across different wavelength ranges. HySens \cite{ref20} proposed a sensor-agnostic hyperspectral foundation model that harmonizes heterogeneous band distributions into a unified spectral grid and uses wavelength-aware embeddings to enhance cross-sensor transferability. HyperFree \cite{ref21} further reduced downstream adaptation costs through a channel-adaptive and tuning-free strategy. In addition, HyperFM \cite{ref24} explored spectral grouping and parameter-efficient modeling for hyperspectral foundation models, showing the potential of group-wise spectral representation learning in large-scale atmospheric retrieval tasks.

Although these methods demonstrate the potential of foundation models for HSI analysis, efficiently adapting them to classification datasets with different spectral band configurations remains challenging. HSIs collected by different sensors often vary in band number, wavelength coverage and band intervals. Existing adaptation strategies commonly rely on spectral compression, band selection, interpolation, spectral harmonization, or fixed-dimensional projection to satisfy the input requirements of pretrained models. These operations improve compatibility but may discard useful spectral information or weaken local spectral continuity. In contrast, this paper focuses on preserving full-band spectral information while adapting hyperspectral foundation models to downstream classification tasks through a multi-branch efficient fine-tuning framework.

\subsection{Parameter-Efficient Fine-Tuning}
Foundation models usually contain hundreds of millions of parameters or more. Fully fine-tuning all parameters for each downstream task requires substantial computational and storage resources. It may also lead to overfitting when only a few labeled HSI samples are available. Parameter-Efficient Fine-Tuning (PEFT) has therefore become an important strategy for adapting large pretrained models \cite{ref25}.

Existing PEFT methods can be broadly divided into adapter-based tuning, prompt-based tuning, and low-rank adaptation \cite{ref26,ref27}. Adapter-based methods insert lightweight trainable modules into a frozen backbone, while prompt-based methods introduce learnable tokens or prompts to guide the pretrained model toward downstream tasks. Low-Rank Adaptation (LoRA) assumes that task-specific weight updates have a low intrinsic rank and approximates the update matrix with two small trainable matrices. Because LoRA freezes the original pretrained weights and only optimizes the low-rank matrices, it can greatly reduce the number of trainable parameters while maintaining adaptation capability.

Recently, PEFT has also been explored in remote sensing foundation models. AiRs \cite{ref28} introduced a remote-sensing-oriented adapter tuning framework by designing spatial context adapters and semantic response adapters, enabling efficient transfer of large pretrained models to object detection, semantic segmentation, and scene classification tasks. UPetu \cite{ref29} further unified adapter tuning and prompt tuning through an efficient quantization adapter module and a context-aware prompt module, reducing the computational and storage costs of adapting remote sensing foundation models. For spectral remote sensing, SpectralX \cite{ref30} proposed a parameter-efficient domain generalization framework that adapts optical remote sensing foundation models to multispectral and hyperspectral modalities through a hyper tokenizer and attribute-oriented adapters. In HSI classification, Ligan et al. \cite{ref31} systematically investigated several PEFT strategies, including LoRA, LoRA+, KronA, LoKr, and KronA+, for adapting a multispectral foundation model to hyperspectral image classification.

\begin{figure*}[!t]
\centering
\includegraphics[width=\textwidth]{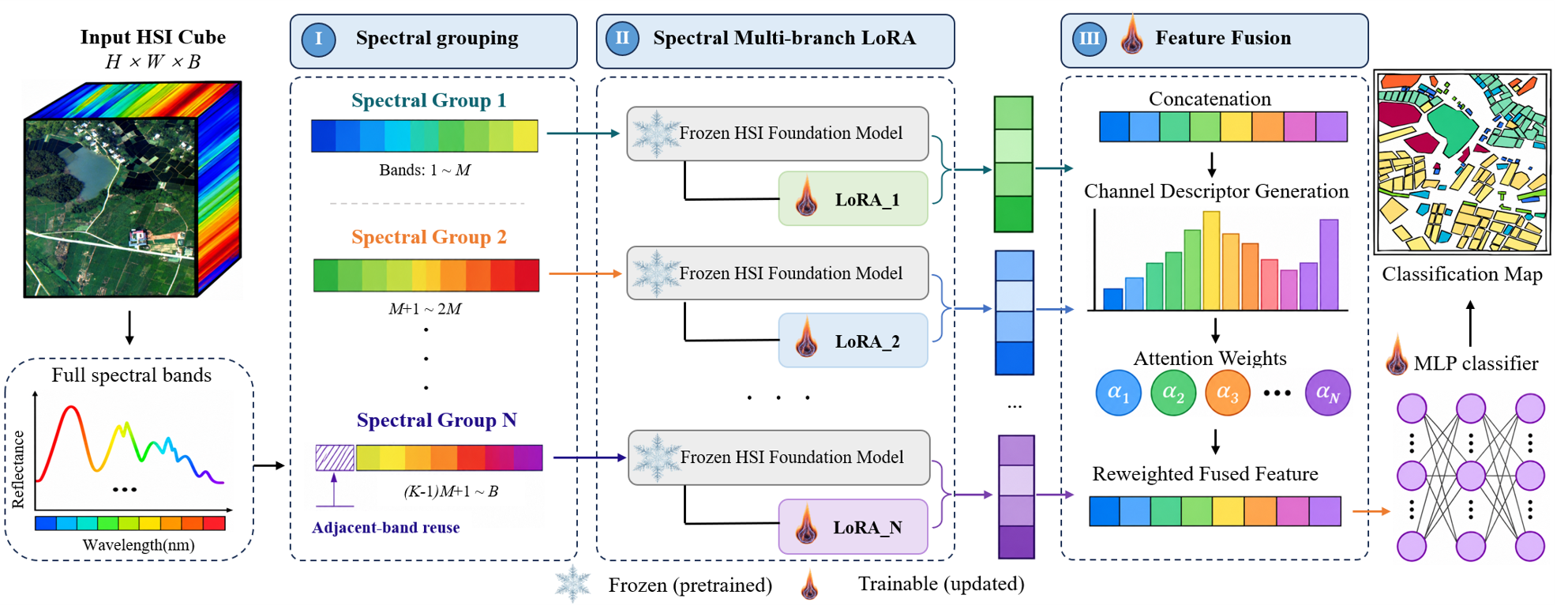}
\caption{Overall architecture of the proposed MBTI framework.}
\label{fig:mbti_architecture}
\end{figure*}

Although these studies demonstrate the effectiveness of PEFT in remote sensing and spectral image interpretation, most existing strategies are designed for a single input stream or general spectral modality adaptation. They do not explicitly consider the continuous full-band structure of HSIs when downstream data are divided to satisfy the input requirements of a pretrained hyperspectral foundation model. In HSI classification, different spectral intervals may contain distinct discriminative information, and applying a single shared adaptation module may be insufficient to model the heterogeneity among band groups. To address this issue, this paper introduces branch-specific LoRA modules into different consecutive spectral branches. In this way, each branch can learn task-specific adaptation for its corresponding spectral interval while the pretrained backbone remains frozen.

\section{Methodology}
The overall workflow of the proposed multi-branch efficient fine-tuning framework for hyperspectral image classification (MBTI) is shown in Fig.~\ref{fig:mbti_architecture}. MBTI consists of three components: 1) spectral-continuity-preserving multi-branch preprocessing, 2) spectral multi-branch LoRA fine-tuning, and 3) multi-branch channel attention fusion.

Given an input HSI with an arbitrary number of spectral bands, MBTI first divides the full-band data into several continuous spectral groups. The grouping follows the input requirement of the pre-trained hyperspectral foundation model. When the last group contains insufficient bands, a band reuse strategy avoids zero padding and preserves local spectral continuity. Each spectral group is then fed into an independent branch built upon the pre-trained foundation model. The backbone, including the base projection, remains frozen, while its LoRA path, the branch-specific Transformer-block LoRA modules, and the lightweight task-specific layers are trained for parameter-efficient task adaptation. Finally, features from all spectral branches are projected, concatenated, and recalibrated by a channel attention fusion module for pixel-wise classification.

\subsection{Multi-Branch Data Preprocessing Based on Spectral Continuity Preservation}
HSIs acquired by different sensors often vary in spectral band number, wavelength coverage, and band intervals. When such data are adapted to a pre-trained foundation model with specific input requirements, direct band compression or selection may discard useful spectral information and disrupt local spectral continuity. To alleviate this problem, we propose a spectral-continuity-preserving multi-branch preprocessing strategy.

\begin{figure}[!t]
\centering
\includegraphics[width=\columnwidth]{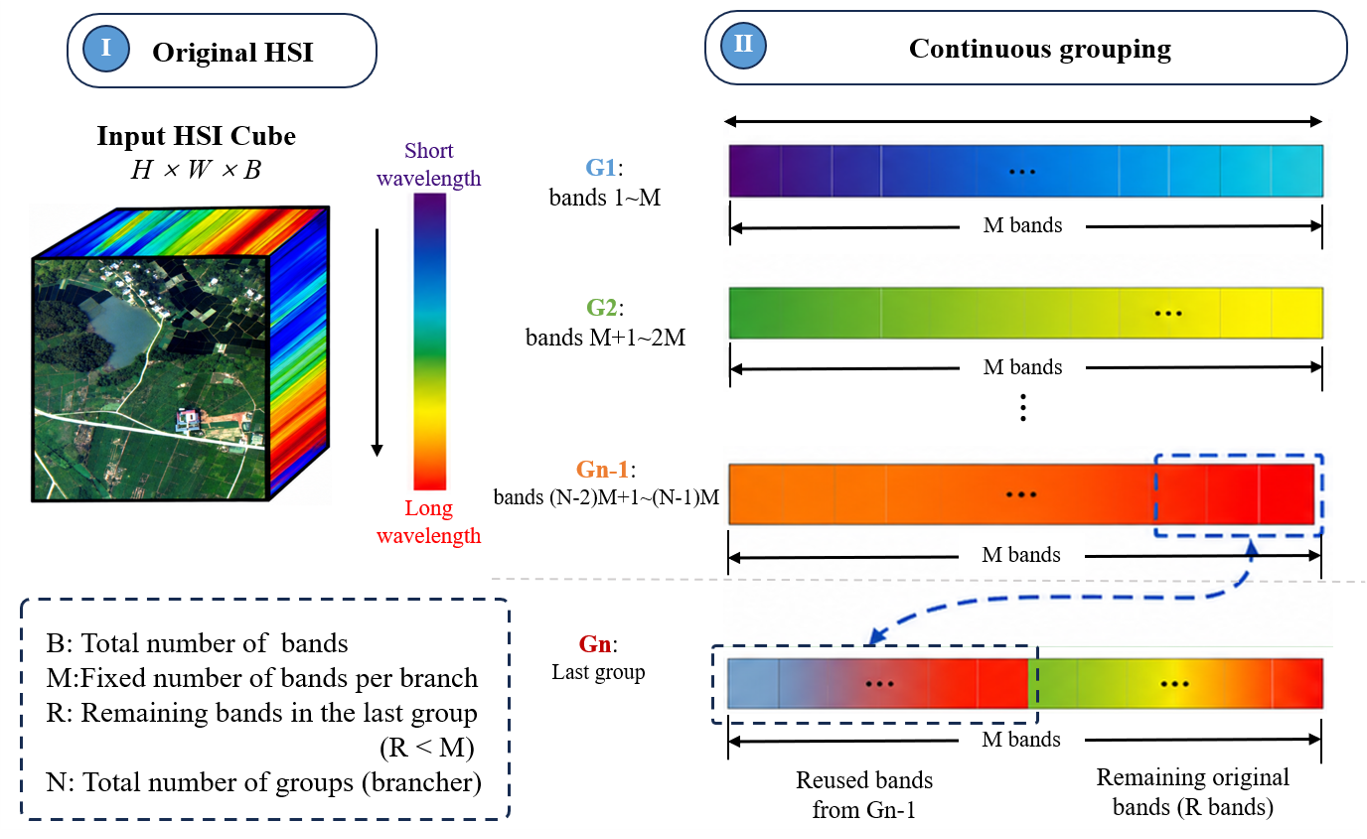}
\caption{Multi-branch Data Preprocessing Based on Spectral Continuity Preservation.}
\label{fig:multi_branch_preprocessing}
\end{figure}

As illustrated in Fig.~\ref{fig:multi_branch_preprocessing}, the proposed strategy divides the original spectrum into continuous groups and constructs the final branch by reusing adjacent bands when necessary. In this way, all branches have the same number of input bands. The original full-band information is retained, and local spectral continuity is preserved as much as possible. For a branch size $M$, each spectral branch uses an $M$-channel patch embedding layer to project its input into 768-dimensional tokens. When $M$ differs from the pretrained channel dimension, the pretrained weights of this layer are not directly loaded. The base projection remains frozen, while its LoRA path, the Transformer-block LoRA modules, and the lightweight task-specific layers are trainable.

Let the input HSI be denoted as
\begin{equation}
\mathbf{X} \in \mathbb{R}^{H \times W \times B},
\end{equation}
where $H$ and $W$ are the spatial height and width, and $B$ is the number of spectral bands. Let $M$ denote the number of bands required by each branch. The number of spectral branches is computed as
\begin{equation}
N = \left\lceil \frac{B}{M} \right\rceil,
\end{equation}
where $\lceil \cdot \rceil$ denotes the ceiling operation.

For the first $N-1$ branches, the input of the $i$-th branch is obtained by selecting a continuous spectral interval:
\begin{equation}
\mathbf{X}_i =
\mathbf{X}_{:,:, (i-1)M+1 : iM},
\quad i=1,2,\ldots,N-1.
\end{equation}
If $B$ is divisible by $M$, the last branch is constructed in the same way. Otherwise, the last branch contains only the remaining spectral bands. Let the number of remaining bands be
\begin{equation}
R = B - (N-1)M,
\quad 0 < R < M.
\end{equation}
Instead of padding zeros, which introduces invalid spectral responses, we reuse adjacent continuous bands from the previous spectral group. The number of reused bands is
\begin{equation}
D = M - R.
\end{equation}
The reused spectral subset is selected from the end of the previous group:
\begin{equation}
\mathbf{X}_{\mathrm{reuse}}
=
\mathbf{X}_{:,:, (N-1)M-D+1 : (N-1)M}.
\end{equation}
The remaining spectral subset is
\begin{equation}
\mathbf{X}_{\mathrm{remain}}
=
\mathbf{X}_{:,:, (N-1)M+1 : B}.
\end{equation}
The input of the last branch is then formed as
\begin{equation}
\mathbf{X}_N =
\mathrm{Concat}_{\lambda}
\left(
\mathbf{X}_{\mathrm{reuse}},
\mathbf{X}_{\mathrm{remain}}
\right),
\end{equation}
where $\mathrm{Concat}_{\lambda}(\cdot)$ denotes concatenation along the spectral dimension. In this way, all branches have the same number of input bands. The original full-band information is retained, and local spectral continuity is preserved as much as possible.

\subsection{Spectral Multi-Branch LoRA Fine-Tuning}
After spectral grouping, each branch processes one continuous spectral subset. This design preserves spectral information, but directly fine-tuning all foundation-model branches would introduce high computational cost. It may also increase the risk of overfitting under limited labeled samples. Therefore, we introduce branch-wise Low-Rank Adaptation (LoRA) for parameter-efficient fine-tuning.

\begin{figure}[!t]
\centering
\includegraphics[width=\columnwidth]{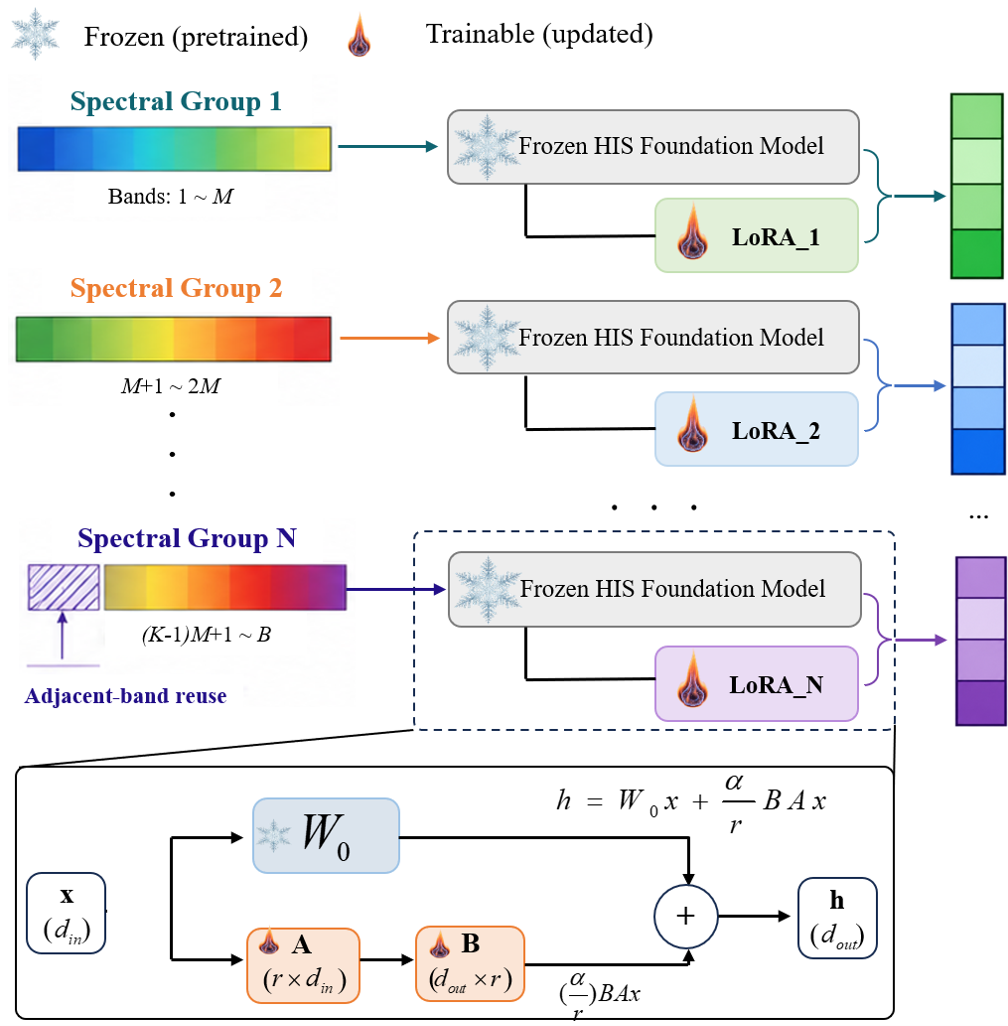}
\caption{Spectral Multi-branch LoRA.}
\label{fig:spectral_multibranch_lora}
\end{figure}

As shown in Fig.~\ref{fig:spectral_multibranch_lora}, LoRA introduces a trainable low-rank path in parallel with the frozen pre-trained linear transformation.

For the $i$-th spectral branch, the grouped input $\mathbf{X}_i$ is fed into a pre-trained encoder equipped with branch-specific LoRA modules:
\begin{equation}
\mathbf{F}_i =
E_i^{\mathrm{LoRA}}(\mathbf{X}_i),
\quad i=1,2,\ldots,N,
\end{equation}
where $E_i^{\mathrm{LoRA}}(\cdot)$ denotes the encoder of the $i$-th branch. The backbone parameters of the encoder are frozen, while the LoRA parameters are trainable and independent across branches. Therefore, LoRA is embedded inside the encoder during feature extraction, rather than being applied after $\mathbf{F}_i$ is obtained.

For a frozen linear transformation in the encoder, let $\mathbf{W}_0$ denote the pre-trained weight matrix and $\mathbf{U}_i$ denote its input feature in the $i$-th branch. The original forward propagation is
\begin{equation}
\mathbf{h}_i =
\mathbf{W}_0 \mathbf{U}_i .
\end{equation}
LoRA introduces a trainable low-rank update:
\begin{equation}
\Delta \mathbf{W}_i =
\frac{\alpha}{r}
\mathbf{B}_i \mathbf{A}_i,
\end{equation}
where $\mathbf{A}_i \in \mathbb{R}^{r \times d_{\mathrm{in}}}$ and $\mathbf{B}_i \in \mathbb{R}^{d_{\mathrm{out}} \times r}$ are the trainable low-rank matrices of the $i$-th branch, $r$ is the LoRA rank, $\alpha$ is a scaling factor, and $\mathbf{W}_0 \in \mathbb{R}^{d_{\mathrm{out}} \times d_{\mathrm{in}}}$. The forward propagation with LoRA is then written as
\begin{equation}
\mathbf{h}_i =
(\mathbf{W}_0+\Delta \mathbf{W}_i)\mathbf{U}_i
=
\mathbf{W}_0\mathbf{U}_i
+
\frac{\alpha}{r}
\mathbf{B}_i\mathbf{A}_i\mathbf{U}_i .
\end{equation}

In our implementation, the above LoRA update is inserted into the linear layers of each Transformer block, including the query, key, value, and output projections in the self-attention module, as well as the two linear transformations in the feed-forward network. In other words, for a target linear layer $\mathbf{W}_t \in \{\mathbf{W}_q,\mathbf{W}_k,\mathbf{W}_v,\mathbf{W}_o,\mathbf{W}_{f1},\mathbf{W}_{f2}\}$, its branch-specific LoRA-adapted form is
\begin{equation}
\mathbf{W}_{t,i}^{\mathrm{LoRA}}
=
\mathbf{W}_t
+
\Delta \mathbf{W}_{t,i},
\quad
t \in \{q,k,v,o,f1,f2\}.
\end{equation}
In this way, each spectral branch learns independent task-specific adaptations for its corresponding spectral interval, while most pre-trained parameters remain frozen. The branch features are then sent to the fusion module described in the next subsection.

\subsection{Multi-Branch Spectral-Channel Recalibration Fusion}
After branch-wise LoRA adaptation, each spectral branch produces a deep feature representation from a specific continuous spectral interval. Directly averaging these branch features is suboptimal because different spectral intervals are not equally informative for different land-cover categories. Simple concatenation may also introduce redundant or weakly relevant responses, especially when adjacent spectral groups are correlated. Therefore, we use a lightweight spectral-channel recalibration fusion module to adaptively integrate branch-wise deep features before classification. The channel reweighting operation follows the global-pooling-and-gating mechanism introduced by squeeze-and-excitation networks~\cite{ref40}, but it is applied to the concatenated multi-branch spectral-channel feature bank rather than to a single-stream feature map.

\begin{figure*}[!t]
\centering
\includegraphics[width=\textwidth]{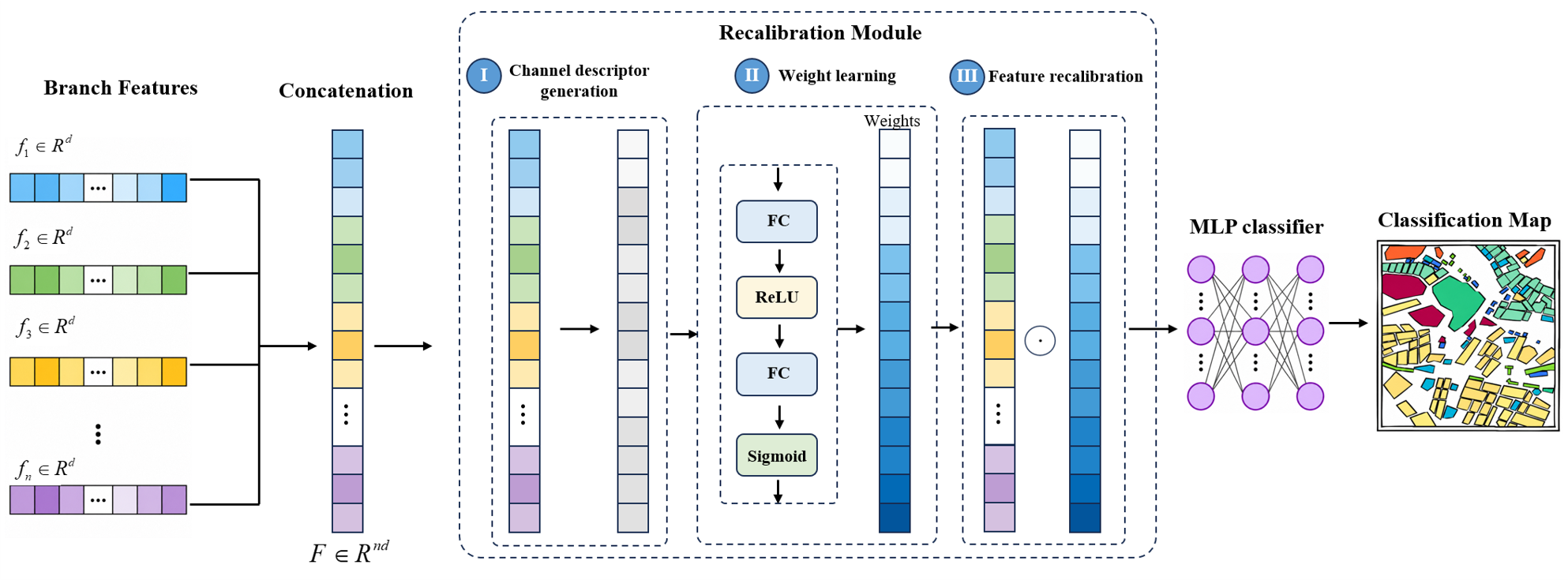}
\caption{Multi-Branch Spectral-Channel Recalibration Fusion.}
\label{fig:spectral_channel_recalibration_fusion}
\end{figure*}

Unlike a generic attention block applied to the final logits, the proposed fusion module operates on classifier-free deep features extracted by each branch. The output of the pre-trained encoder is first compressed by a branch-specific projection layer. For the $i$-th branch, the projected feature is formulated as
\begin{equation}
\mathbf{Z}_i = P_i(E_i^{\mathrm{LoRA}}(\mathbf{X}_i)),
\quad i=1,2,\ldots,N,
\end{equation}
where $E_i^{\mathrm{LoRA}}(\cdot)$ denotes the frozen foundation-model encoder equipped with branch-specific LoRA, $P_i(\cdot)$ is a trainable projection layer without LoRA, and $\mathbf{Z}_i \in \mathbb{R}^{L \times C}$ is the projected branch feature. Here, $L$ denotes the token length of the branch feature sequence, and $C$ is the unified feature dimension of each branch. In our implementation, $P_i(\cdot)$ compresses the concatenated multi-level encoder features into a compact branch representation. This reduces the cost of subsequent fusion while preserving discriminative information before classification.

\begin{algorithm}[!t]
\caption{Multi-Branch Efficient Fine-Tuning Framework for Hyperspectral Image Classification}
\label{alg:mbti_fusion}
\begin{algorithmic}[1]
\REQUIRE Branch inputs $\{\mathbf{X}_i\}_{i=1}^{N}$; frozen encoders equipped with branch-wise LoRA $\{E_i^{\mathrm{LoRA}}\}_{i=1}^{N}$; projection layers $\{P_i\}_{i=1}^{N}$.
\ENSURE Classification logits $\mathbf{O}$.
\FOR{each branch $i = 1,\ldots,N$}
\STATE Extract branch feature: $\mathbf{F}_i = E_i^{\mathrm{LoRA}}(\mathbf{X}_i)$.
\STATE Project deep feature: $\mathbf{Z}_i = P_i(\mathbf{F}_i)$.
\ENDFOR
\STATE Concatenate branch features: $\mathbf{Z} = \mathrm{Concat}_{c}(\mathbf{Z}_1,\ldots,\mathbf{Z}_N)$.
\STATE Compute global spectral-channel descriptor: $s_c = \frac{1}{L}\sum_{l=1}^{L}Z_{l,c}$.
\STATE Generate calibration weights: $\mathbf{g} = \sigma(\mathbf{W}_2\delta(\mathbf{W}_1\mathbf{s}))$.
\STATE Recalibrate fused feature: $\widetilde{\mathbf{Z}}_{l,c} = g_c \mathbf{Z}_{l,c}$.
\STATE Obtain compact discriminative feature: $\mathbf{H} = \mathrm{MLP}(\widetilde{\mathbf{Z}})$.
\STATE Predict logits: $\mathbf{O} = \mathrm{Classifier}(\mathbf{H})$.
\RETURN $\mathbf{O}$.
\end{algorithmic}
\end{algorithm}

The projected branch features are then concatenated along the channel dimension:
\begin{equation}
\mathbf{Z}
=
\mathrm{Concat}_{c}
\left(
\mathbf{Z}_1,\mathbf{Z}_2,\ldots,\mathbf{Z}_N
\right),
\quad
\mathbf{Z} \in \mathbb{R}^{L \times NC}.
\end{equation}
This concatenated representation can be viewed as a spectral-channel feature bank. Each channel corresponds to a feature response from a spectral branch. Instead of treating all branch channels equally, we learn a global calibration vector from the fused representation. First, global information is aggregated over the token dimension:
\begin{equation}
s_c =
\frac{1}{L}
\sum_{l=1}^{L}
Z_{l,c},
\quad c=1,2,\ldots,NC,
\end{equation}
where $s_c$ summarizes the global response of the $c$-th spectral-channel feature. The channel descriptor $\mathbf{s}$ is then passed through a lightweight bottleneck gating function:
\begin{equation}
\mathbf{g}
=
\sigma
\left(
\mathbf{W}_2
\delta
\left(
\mathbf{W}_1 \mathbf{s}
\right)
\right),
\quad
\mathbf{g} \in \mathbb{R}^{NC},
\end{equation}
where $\mathbf{W}_1 \in \mathbb{R}^{\frac{NC}{\rho} \times NC}$ and $\mathbf{W}_2 \in \mathbb{R}^{NC \times \frac{NC}{\rho}}$ are learnable transformation matrices, $\rho$ is the reduction ratio, $\delta(\cdot)$ is the ReLU activation, and $\sigma(\cdot)$ is the sigmoid function. The learned vector $\mathbf{g}$ estimates the importance of all branch-channel responses. Since the channels are organized by spectral branches, this operation models both inter-branch importance and intra-branch channel relevance.

The original concatenated feature is recalibrated by the learned spectral-channel weights:
\begin{equation}
\widetilde{\mathbf{Z}}_{l,c}
=
g_c \mathbf{Z}_{l,c},
\quad
l=1,2,\ldots,L,\; c=1,2,\ldots,NC.
\end{equation}
 This operation emphasizes informative spectral-channel responses and suppresses redundant or weakly relevant ones. The learned vector $\mathbf{g}$ estimates the importance of all branch-channel responses. Since the channels are organized by spectral branches, this operation models both inter-branch importance and intra-branch channel relevance. Therefore, the learned weights jointly reflect inter-branch spectral importance and intra-branch feature relevance before the final classifier. The recalibrated feature is then fed into a progressive transition head:
\begin{equation}
\mathbf{H}
=
\phi_3
\left(
\phi_2
\left(
\phi_1
\left(
\widetilde{\mathbf{Z}}\mathbf{W}_1' + \mathbf{b}_1'
\right)
\mathbf{W}_2' + \mathbf{b}_2'
\right)
\mathbf{W}_3 + \mathbf{b}_3
\right),
\end{equation}
where $\mathbf{W}_1' \in \mathbb{R}^{NC \times d_1}$, $\mathbf{W}_2' \in \mathbb{R}^{d_1 \times d_2}$, and $\mathbf{W}_3 \in \mathbb{R}^{d_2 \times d_h}$ are learnable linear transformations in the transition head. $\mathbf{b}_1'$, $\mathbf{b}_2'$, and $\mathbf{b}_3$ are bias terms, and $\phi_1(\cdot)$, $\phi_2(\cdot)$, and $\phi_3(\cdot)$ denote activation and dropout operations. The transition head maps the high-dimensional multi-branch feature into a compact discriminative representation. Finally, a linear classifier produces the class logits:
\begin{equation}
\mathbf{O}
=
\mathbf{H}\mathbf{W}_{\mathrm{cls}}
+
\mathbf{b}_{\mathrm{cls}},
\end{equation}
where $\mathbf{W}_{\mathrm{cls}} \in \mathbb{R}^{d_h \times K}$ and $\mathbf{b}_{\mathrm{cls}} \in \mathbb{R}^{K}$, with $K$ denoting the number of land-cover classes. The fusion module is trained jointly with branch-wise LoRA and projection layers, while the original pre-trained backbone remains frozen. Therefore, the proposed fusion design integrates information from different spectral groups and learns how much each branch-channel response should contribute to classification.  The overall procedure of MBTI is summarized in Algorithm~\ref{alg:mbti_fusion}.

\section{Experiments and Analysis}
\subsection{Datasets}
\begin{figure}[!t]
\centering
\includegraphics[width=\columnwidth]{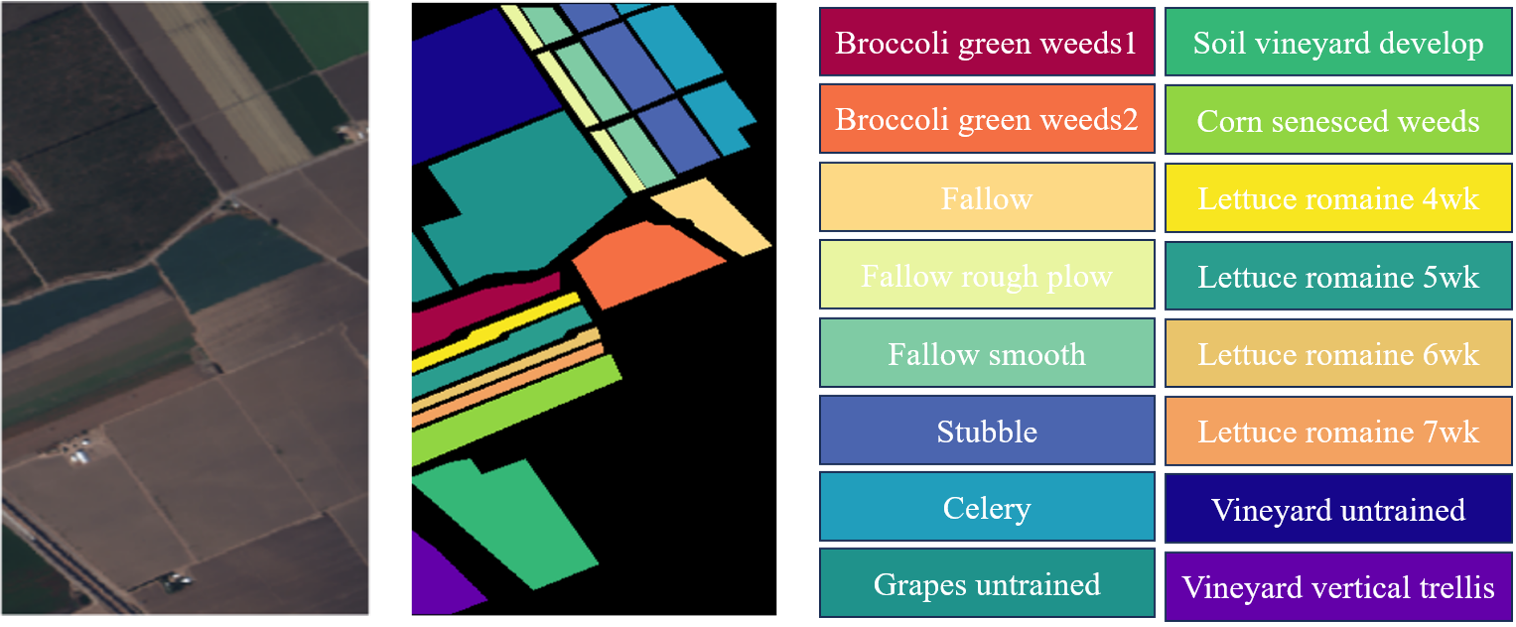}
\caption{Description of the Salinas.}
\label{fig:salinas_dataset}
\end{figure}

\begin{figure}[!t]
\centering
\includegraphics[width=\columnwidth]{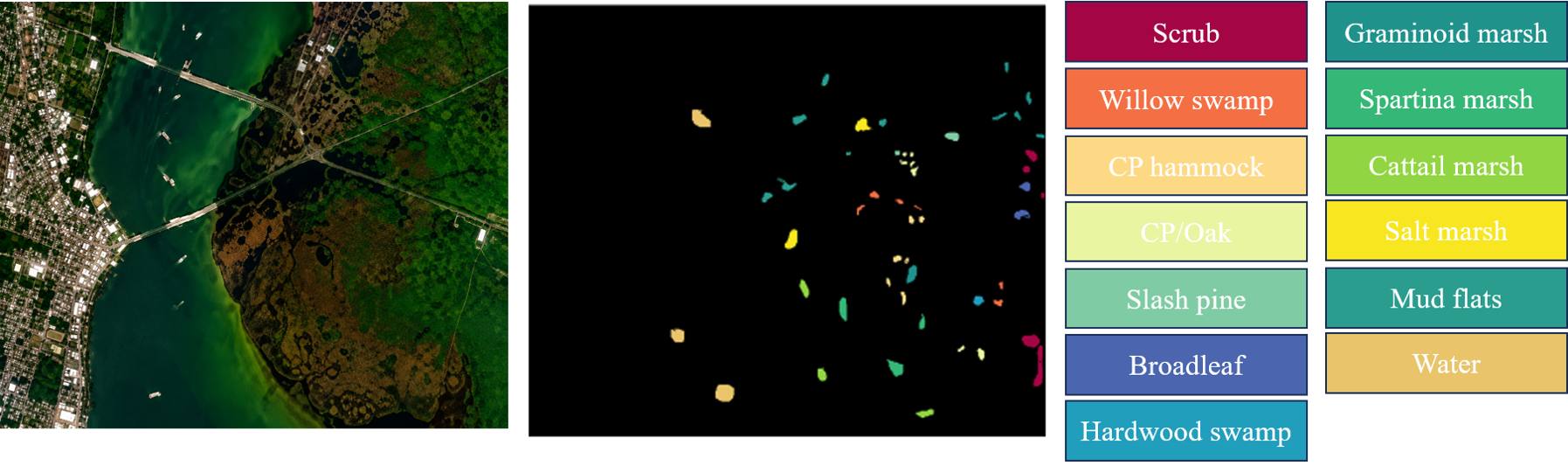}
\caption{Description of the Kennedy Space Center.}
\label{fig:ksc_dataset}
\end{figure}

\begin{figure}[!t]
\centering
\includegraphics[width=\columnwidth]{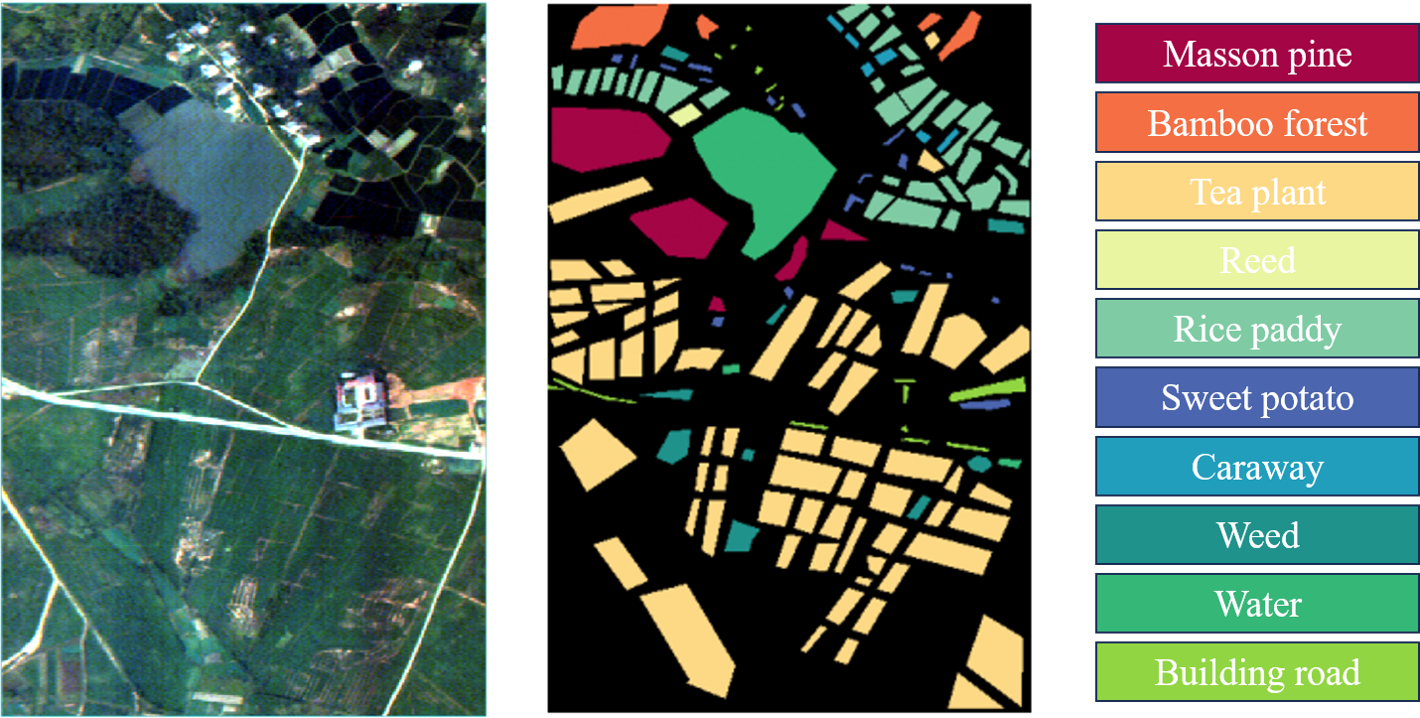}
\caption{Description of the FangluTeaFarm.}
\label{fig:tf_dataset}
\end{figure}

To comprehensively evaluate MBTI, experiments were conducted on three widely used HSI classification datasets: Salinas (SA), Kennedy Space Center (KSC), and FangluTeaFarm (TF). These datasets were acquired by different sensors and cover diverse spatial resolutions, spectral ranges, scene types, and land-cover categories. They therefore provide a representative benchmark for evaluating the proposed method. Dataset visualizations and class statistics are provided in Figs.~\ref{fig:salinas_dataset}--\ref{fig:tf_dataset} and Tables~I--III.

\textit{Salinas dataset (SA).} The SA dataset was collected by the AVIRIS sensor over agricultural areas in Salinas Valley, California. As shown in Fig.~\ref{fig:salinas_dataset} and Table~I, it contains 224 spectral bands covering 400--2500 nm and has a spatial resolution of 3.7 m. After removing 20 water absorption bands, 204 bands are retained for experiments. The dataset contains 16 land-cover classes, mainly corresponding to crop types such as stubble, celery, and lettuce. Because of its large number of categories and notable intra-class variation, Salinas is widely used as a challenging benchmark for HSI classification.

\textit{Kennedy Space Center dataset (KSC).} The KSC dataset was acquired by the airborne AVIRIS sensor over the Kennedy Space Center area, Florida, USA, in 1996. As shown in Fig.~\ref{fig:ksc_dataset} and Table~II, the original scene contains 224 spectral bands and has a spatial resolution of 18 m. Following the commonly used preprocessing protocol, water absorption and noisy bands were removed, and 176 spectral bands were retained in our experiments. The scene includes complex ecological environments and diverse vegetation communities. It contains 13 land-cover classes, such as Scrub, Cabbage palm hammock, Graminoid marsh, and Mud flats. Due to subtle spectral differences among vegetation categories and complex spatial distributions, KSC is commonly used to evaluate the robustness of HSI classification methods.

\textit{FangluTeaFarm dataset (TF).} The TF dataset was captured in China in 2017 using the Pushbroom Hyperspectral Imager (PHI). As shown in Fig.~\ref{fig:tf_dataset} and Table~III, the image size is $348 \times 512$ pixels, with a spatial resolution of 2.25 m. The dataset contains 80 spectral bands ranging from 417 nm to 855 nm. It represents a tea-farm scene with 10 land-cover categories, including Masson pine, bamboo forest, tea plant, rice paddy, water, and building road. This dataset provides a practical benchmark for complex agricultural and forestry environments.

\begin{table}[!t]
\centering
\footnotesize
\caption{Categories and sample size of the Salinas dataset.}
\label{tab:salinas_categories}
\begin{tabular}{c l r}
\hline
Class & Name & Sample size \\
\hline
1  & Broccoli green weeds1        & 2009  \\
2  & Broccoli green weeds2        & 3726  \\
3  & Fallow                      & 1976  \\
4  & Fallow rough plow           & 1394  \\
5  & Fallow smooth               & 2678  \\
6  & Stubble                     & 3959  \\
7  & Celery                      & 3579  \\
8  & Grapes untrained            & 11271 \\
9  & Soil vineyard develop       & 6203  \\
10 & Corn senesced weeds        & 3278  \\
11 & Lettuce romaine 4wk         & 1068  \\
12 & Lettuce romaine 5wk         & 1927  \\
13 & Lettuce romaine 6wk         & 916   \\
14 & Lettuce romaine 7wk         & 1070  \\
15 & Vineyard untrained           & 7268  \\
16 & Vineyard vertical trellis    & 1807  \\
\hline
\multicolumn{2}{c}{Total} & 54129 \\
\hline
\end{tabular}
\end{table}

\begin{table}[!t]
\centering
\footnotesize
\caption{Categories and sample size of the KSC dataset.}
\label{tab:ksc_categories}
\begin{tabular}{c l r}
\hline
Class & Name & Sample size \\
\hline
1  & Scrub             & 761 \\
2  & Willow swamp      & 243 \\
3  & Cabbage palm hammock        & 256 \\
4  & Cabbage palm/Oak            & 252 \\
5  & Slash pine        & 161 \\
6  & Oak/Broadleaf     & 229 \\
7  & Hardwood swamp    & 105 \\
8  & Graminoid marsh   & 431 \\
9  & Spartina marsh    & 520 \\
10 & Cattail marsh     & 404 \\
11 & Salt marsh        & 419 \\
12 & Mud flats         & 503 \\
13 & Water             & 927 \\
\hline
\multicolumn{2}{c}{Total} & 5211 \\
\hline
\end{tabular}
\end{table}

\begin{table}[!t]
\centering
\footnotesize
\caption{Categories and sample size of the FangluTeaFarm dataset.}
\label{tab:tf_categories}
\begin{tabular}{c l r}
\hline
Class & Name & Sample size \\
\hline
1  & Masson pine    & 5806  \\
2  & Bamboo forest  & 2318  \\
3  & Tea plant      & 28428 \\
4  & Reed           & 214   \\
5  & Rice paddy     & 6809  \\
6  & Sweet potato   & 817   \\
7  & Caraway        & 429   \\
8  & Weed           & 1861  \\
9  & Water          & 6141  \\
10 & Building road  & 911   \\
\hline
\multicolumn{2}{c}{Total} & 53734 \\
\hline
\end{tabular}
\end{table}

\subsection{Experimental Settings}
\subsubsection{Evaluation Metrics}
To quantitatively evaluate classification performance, four commonly used metrics were adopted: overall accuracy (OA), class-wise accuracy, average accuracy (AA), and Cohen's Kappa coefficient. OA measures the proportion of correctly classified samples:
\begin{equation}
\mathrm{OA}
=
\frac{\sum_{i=1}^{K} n_{ii}}{N},
\end{equation}
where $K$ is the number of classes, $N$ is the total number of test samples, and $n_{ii}$ denotes the number of correctly classified samples in the $i$-th class. The class-wise accuracy of the $i$-th class is defined as
\begin{equation}
\mathrm{Acc}_i
=
\frac{n_{ii}}{n_{i+}},
\end{equation}
where $n_{i+}$ denotes the total number of samples belonging to the $i$-th ground-truth class. AA is computed as the mean of class-wise accuracies:
\begin{equation}
\mathrm{AA}
=
\frac{1}{K}
\sum_{i=1}^{K}
\mathrm{Acc}_i.
\end{equation}
Cohen's Kappa coefficient evaluates the agreement between predictions and ground-truth labels while considering chance agreement:
\begin{equation}
\kappa
=
\frac{
N\sum_{i=1}^{K} n_{ii}
-
\sum_{i=1}^{K} n_{i+}n_{+i}
}{
N^2
-
\sum_{i=1}^{K} n_{i+}n_{+i}
},
\end{equation}
where $n_{+i}$ denotes the number of samples predicted as the $i$-th class.

\subsubsection{Hyperspectral Foundation Model}
HyperSIGMA was selected as the pre-trained hyperspectral foundation model. This choice is based on its strong representation capability for hyperspectral imagery and its demonstrated effectiveness on multiple downstream HSI tasks. Unlike general remote sensing foundation models or multispectral-oriented models, HyperSIGMA is designed for hyperspectral interpretation and learns representations from continuous spectral intervals. This property is consistent with the spectral grouping strategy adopted in MBTI. We used the publicly available parameters of its base version to initialize the spatial encoder in each branch. During fine-tuning, the pre-trained backbone parameters were frozen, and only the branch-wise LoRA modules, feature projection layers, and fusion classifier were updated.

\subsubsection{Implementation Details}
For a fair comparison, all methods followed the same training protocol. For each dataset, 10 labeled samples per class were randomly selected for training. After training, MBTI and its controlled variants were evaluated on all labeled pixels to obtain stable full-scene accuracy statistics. The main training and adaptation configurations are summarized as follows. The input image was divided into $128\times128$ spatial patches, and the ViT patch size was set to 8. The model was optimized using AdamW with a learning rate of 6e-5, betas of (0.9, 0.999), and a weight decay of 0.05. A cosine annealing learning-rate scheduler was adopted.

\subsection{Algorithm Performance and State-of-the-Art Comparisons}
To evaluate the effectiveness of MBTI, we compare it with representative methods from several categories. These include classical machine learning, supervised deep learning, cross-domain few-shot learning, hyperspectral foundation models, and fine-tuning-free large models. Specifically, the compared methods include SVM \cite{ref32}; five supervised deep learning methods, i.e., CDCNN \cite{ref33}, DBDA \cite{ref34}, DBMA \cite{ref35}, FDSSC \cite{ref36}, and SSRN \cite{ref7}; three cross-domain few-shot learning methods, i.e., DCFSL \cite{ref37}, Gia-CFSL \cite{ref38}, and HFSL \cite{ref39}; the hyperspectral foundation model HyperSIGMA \cite{ref18}; and the fine-tuning-free hyperspectral foundation model HyperFree \cite{ref21}. Experiments are conducted on KSC, Salinas, and FangluTeaFarm. Classification performance is evaluated using OA, AA, and Kappa. For clarity, all Kappa values reported in the tables are multiplied by 100.

\begin{table*}[!t]
\centering
\footnotesize
\caption{Classification results (\%) on the KSC dataset with different methods.}
\label{tab:ksc_results}
\resizebox{\textwidth}{!}{
\begin{tabular}{l c c c c c c c c c c c c}
\hline
\textbf{Class} & \textbf{SVM} & \textbf{CDCNN} & \textbf{DBDA} & \textbf{DBMA} & \textbf{FDSSC} & \textbf{SSRN} & \textbf{DCFSL} & \textbf{HFSL} & \textbf{Gia-CFSL} & \textbf{HyperFree} & \textbf{HyperSIGMA} & \textbf{MBTI} \\
\hline
Scrub            & 88.28 & 94.31 & 98.91 & 98.26 & 99.45 & 98.77 & 98.67 & 94.41 & 99.47 & 0.00  & 95.74 & 96.19 \\
Willow swamp     & 71.24 & 92.13 & 96.40 & 92.43 & 97.30 & 97.83 & 90.13 & 98.28 & 82.83 & 0.00  & 93.99 & 100.00 \\
CP hammock       & 88.21 & 55.69 & 95.58 & 100.00 & 66.95 & 99.44 & 96.34 & 98.37 & 93.90 & 0.00  & 100.00 & 98.83 \\
CP/Oak           & 67.77 & 51.72 & 85.83 & 64.62 & 79.55 & 76.84 & 74.38 & 97.52 & 78.93 & 1.59  & 87.60 & 95.63 \\
Slash pine       & 55.63 & 56.06 & 90.63 & 65.71 & 91.41 & 87.50 & 81.46 & 98.01 & 82.12 & 0.00  & 100.00 & 100.00 \\
Oak/Broadleaf    & 59.36 & 68.63 & 85.96 & 97.78 & 99.05 & 96.31 & 97.72 & 100.00 & 99.09 & 0.00  & 100.00 & 100.00 \\
Hardwood swamp   & 97.89 & 80.37 & 100.00 & 98.75 & 94.51 & 100.00 & 100.00 & 100.00 & 100.00 & 100.00 & 100.00 & 100.00 \\
Graminoid marsh  & 81.47 & 66.92 & 86.32 & 80.54 & 93.36 & 92.69 & 100.00 & 97.15 & 99.76 & 58.24 & 77.67 & 100.00 \\
Spartina marsh   & 88.82 & 72.81 & 100.00 & 94.29 & 99.18 & 96.68 & 100.00 & 86.67 & 100.00 & 37.12 & 100.00 & 100.00 \\
Cattail marsh    & 81.22 & 97.59 & 100.00 & 94.26 & 100.00 & 96.01 & 89.09 & 100.00 & 91.37 & 72.03 & 100.00 & 99.75 \\
Salt marsh       & 98.29 & 88.67 & 100.00 & 89.58 & 100.00 & 100.00 & 98.29 & 100.00 & 97.80 & 0.00  & 100.00 & 100.00 \\
Mud flats        & 81.54 & 92.89 & 98.96 & 96.70 & 99.17 & 100.00 & 99.19 & 99.59 & 99.59 & 79.52 & 100.00 & 100.00 \\
Water            & 99.02 & 98.78 & 100.00 & 100.00 & 100.00 & 100.00 & 100.00 & 100.00 & 100.00 & 99.68 & 100.00 & 100.00 \\
\hline
\textbf{OA}     & 85.51 & 84.10 & 96.47 & 92.06 & 95.78 & 96.57 & 96.24 & 97.22 & 96.36 & 41.60 & 96.65 & \textbf{99.16} \\
\textbf{AA}     & 81.44 & 78.20 & 95.25 & 90.22 & 93.84 & 95.54 & 94.25 & 97.69 & 94.22 & 34.47 & 96.54 & \textbf{99.26} \\
\textbf{Kappa}  & 83.89 & 82.27 & 96.06 & 91.14 & 95.29 & 96.17 & 95.81 & 96.91 & 95.94 & 36.06 & 96.27 & \textbf{99.06} \\
\hline
\end{tabular}}
\end{table*}

\begin{table*}[!t]
\centering
\footnotesize
\caption{Classification results (\%) on the Salinas dataset with different methods.}
\label{tab:salinas_results}
\resizebox{\textwidth}{!}{
\begin{tabular}{l c c c c c c c c c c c c}
\hline
\textbf{Class} & \textbf{SVM} & \textbf{CDCNN} & \textbf{DBDA} & \textbf{DBMA} & \textbf{FDSSC} & \textbf{SSRN} & \textbf{DCFSL} & \textbf{HFSL} & \textbf{Gia-CFSL} & \textbf{HyperFree} & \textbf{HyperSIGMA} & \textbf{MBTI} \\
\hline
Broccoli green weeds1     & 96.35 & 79.57 & 100.00 & 100.00 & 100.00 & 100.00 & 100.00 & 100.00 & 99.95 & 99.40 & 99.90 & 96.62 \\
Broccoli green weeds2     & 97.85 & 96.55 & 100.00 & 100.00 & 100.00 & 100.00 & 99.97 & 98.30 & 100.00 & 99.84 & 99.92 & 99.57 \\
Fallow                    & 99.24 & 95.14 & 96.56 & 97.71 & 97.61 & 91.04 & 99.95 & 100.00 & 99.95 & 100.00 & 100.00 & 100.00 \\
Fallow rough plow         & 99.49 & 95.76 & 94.24 & 87.27 & 94.78 & 96.64 & 99.86 & 100.00 & 99.57 & 20.30 & 90.90 & 91.39 \\
Fallow smooth             & 91.38 & 99.80 & 99.61 & 99.83 & 99.92 & 99.77 & 91.12 & 96.70 & 95.16 & 98.78 & 80.32 & 92.64 \\
Stubble                   & 97.87 & 99.34 & 100.00 & 100.00 & 99.97 & 99.92 & 100.00 & 99.65 & 100.00 & 58.31 & 87.77 & 94.75 \\
Celery                    & 98.94 & 97.01 & 99.97 & 99.92 & 100.00 & 99.92 & 98.35 & 99.94 & 95.83 & 99.97 & 98.71 & 98.66 \\
Grapes untrained          & 50.91 & 71.10 & 91.64 & 87.19 & 85.75 & 84.91 & 81.27 & 87.35 & 63.49 & 95.57 & 96.65 & 94.10 \\
Soil vineyard develop     & 98.60 & 98.13 & 99.34 & 99.40 & 99.37 & 99.28 & 98.66 & 99.39 & 96.53 & 99.90 & 99.48 & 99.24 \\
Corns senesced weeds      & 85.16 & 87.01 & 96.26 & 96.58 & 98.43 & 97.33 & 79.68 & 95.72 & 79.35 & 43.84 & 99.91 & 97.96 \\
Lettuce romaine 4wk       & 92.53 & 68.25 & 94.61 & 87.25 & 92.69 & 92.36 & 97.83 & 99.53 & 99.72 & 96.44 & 98.58 & 95.51 \\
Lettuce romaine 5wk       & 99.48 & 88.09 & 96.07 & 100.00 & 98.15 & 99.58 & 100.00 & 99.48 & 100.00 & 53.40 & 92.18 & 98.86 \\
Lettuce romaine 6wk       & 99.01 & 86.29 & 99.78 & 98.47 & 97.94 & 96.37 & 99.45 & 100.00 & 99.78 & 0.00 & 100.00 & 100.00 \\
Lettuce romaine 7wk       & 87.74 & 92.88 & 81.11 & 95.35 & 79.37 & 82.85 & 96.13 & 98.96 & 95.94 & 95.51 & 93.02 & 98.97 \\
Vineyard untrained        & 62.46 & 52.87 & 74.28 & 82.16 & 84.56 & 75.99 & 65.32 & 85.15 & 74.08 & 99.44 & 98.31 & 96.13 \\
Vineyard vertical trellis & 92.04 & 72.85 & 100.00 & 95.69 & 100.00 & 99.78 & 95.27 & 100.00 & 97.89 & 100.00 & 100.00 & 100.00 \\
\hline
\textbf{OA}    & 81.98 & 81.99 & 93.06 & 93.55 & 93.84 & 92.30 & 89.20 & 94.67 & 86.56 & 86.38 & 96.37 & \textbf{96.68} \\
\textbf{AA}    & 90.56 & 86.29 & 95.22 & 95.43 & 95.54 & 94.73 & 93.93 & 97.51 & 93.58 & 78.80 & 95.98 & 97.15 \\
\textbf{Kappa} & 80.04 & 80.06 & 92.29 & 92.82 & 93.13 & 91.43 & 87.97 & 94.07 & 85.09 & 84.88 & 95.96 & \textbf{96.31} \\
\hline
\end{tabular}}
\end{table*}

\begin{table*}[!t]
\centering
\footnotesize
\caption{Classification results (\%) on the FangluTeaFarm dataset with different methods.}
\label{tab:teafarm_results}
\resizebox{\textwidth}{!}{
\begin{tabular}{l c c c c c c c c c c c c}
\hline
\textbf{Class} & \textbf{SVM} & \textbf{CDCNN} & \textbf{DBDA} & \textbf{DBMA} & \textbf{FDSSC} & \textbf{SSRN} & \textbf{DCFSL} & \textbf{HFSL} & \textbf{Gia-CFSL} & \textbf{HyperFree} & \textbf{HyperSIGMA} & \textbf{MBTI} \\
\hline
Masson pine     & 76.92 & 94.04 & 89.88 & 91.14 & 89.53 & 93.50 & 99.52 & 97.86 & 99.50 & 89.67 & 91.70 & 97.00 \\
Bamboo forest   & 51.04 & 26.41 & 46.30 & 42.16 & 49.78 & 46.18 & 76.99 & 78.42 & 70.02 & 100.00 & 99.05 & 94.52 \\
Tea plant       & 72.89 & 99.39 & 99.61 & 99.03 & 99.58 & 98.93 & 89.90 & 90.20 & 90.39 & 96.43 & 85.01 & 95.93 \\
Reed            & 89.22 & 70.23 & 66.23 & 75.28 & 75.84 & 80.00 & 100.00 & 100.00 & 95.59 & 100.00 & 100.00 & 100.00 \\
Rice paddy      & 97.66 & 85.88 & 99.34 & 98.63 & 99.05 & 99.21 & 99.60 & 99.97 & 99.76 & 98.15 & 92.69 & 93.93 \\
Sweet potato    & 86.62 & 22.20 & 73.45 & 91.50 & 27.98 & 85.51 & 99.01 & 99.63 & 99.50 & 63.58 & 55.27 & 86.66 \\
Caraway         & 94.27 & 61.27 & 82.16 & 84.82 & 80.73 & 82.32 & 100.00 & 100.00 & 99.28 & 100.00 & 100.00 & 100.00 \\
Weed            & 67.80 & 47.87 & 47.67 & 52.51 & 41.42 & 48.35 & 75.80 & 82.93 & 84.12 & 59.00 & 90.65 & 84.85 \\
Water           & 98.70 & 100.00 & 100.00 & 100.00 & 100.00 & 100.00 & 98.83 & 97.60 & 99.49 & 97.88 & 96.95 & 99.80 \\
Building road   & 96.23 & 77.85 & 96.96 & 95.18 & 97.71 & 99.44 & 100.00 & 100.00 & 95.23 & 92.41 & 80.80 & 93.19 \\
\hline
\textbf{OA}    & 79.13 & 80.89 & 90.96 & 91.29 & 86.99 & 91.54 & 92.57 & 92.78 & 92.82 & 92.86 & 88.53 & \textbf{95.65} \\
\textbf{AA}    & 83.13 & 68.51 & 80.16 & 83.03 & 76.16 & 83.34 & 93.69 & \textbf{94.66} & 93.29 & 89.71 & 89.21 & 94.59 \\
\textbf{Kappa} & 71.96 & 74.33 & 87.21 & 87.61 & 82.15 & 87.95 & 89.40 & 89.67 & 89.72 & 89.57 & 83.90 & \textbf{93.66} \\
\hline
\end{tabular}}
\end{table*}

\begin{figure}[!t]
\centering
\includegraphics[width=\columnwidth]{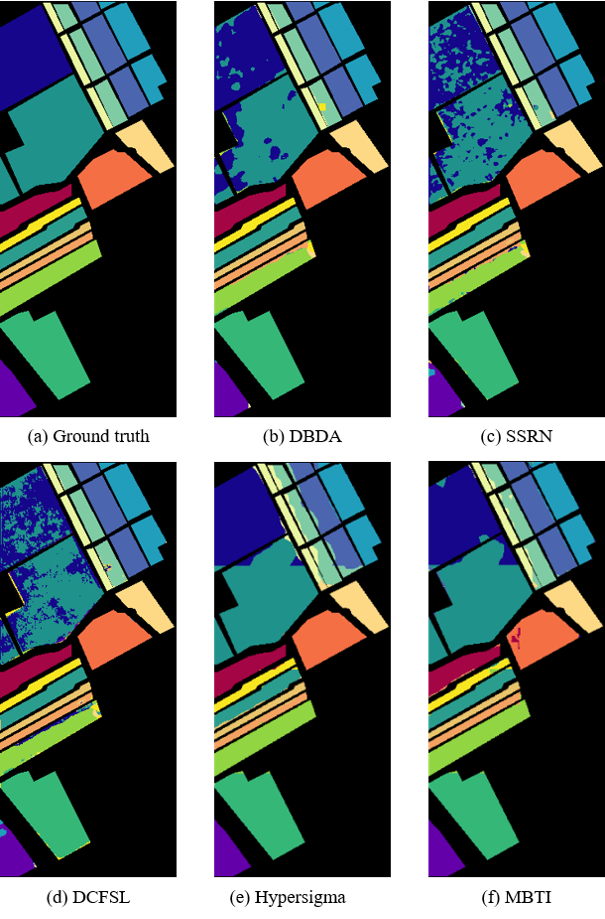}
\caption{Classification maps for Salinas obtained with different methods.}
\label{fig:salinas_classification_maps}
\end{figure}

\begin{figure}[!t]
\centering
\includegraphics[width=\columnwidth]{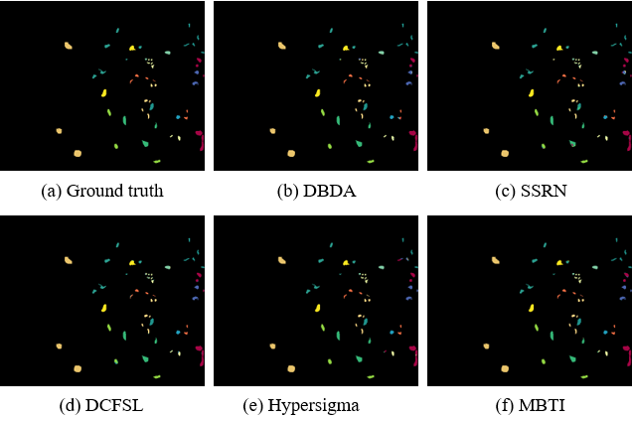}
\caption{Classification maps for KSC obtained with different methods.}
\label{fig:ksc_classification_maps}
\end{figure}

\begin{figure}[!t]
\centering
\includegraphics[width=\columnwidth]{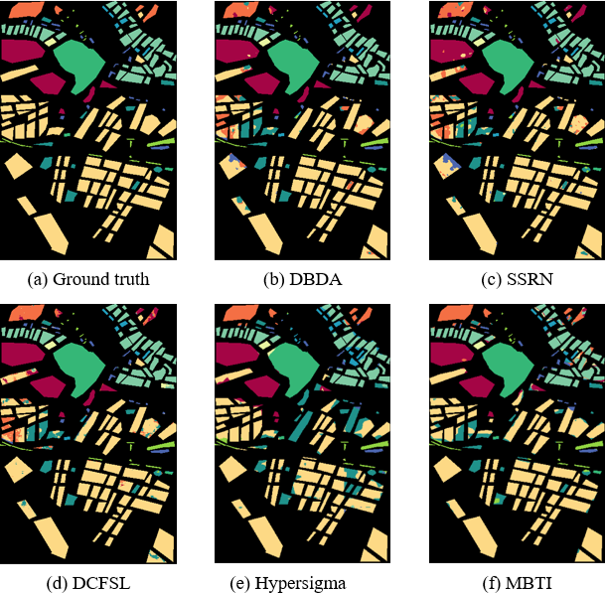}
\caption{Classification maps for FangluTeaFarm obtained with different methods.}
\label{fig:tf_classification_maps}
\end{figure}

As shown in Tables~\ref{tab:ksc_results}--\ref{tab:teafarm_results}, MBTI achieves the best OA and Kappa on all three datasets. On KSC, MBTI obtains 99.16\% OA, 99.26\% AA, and 99.06\% Kappa. It outperforms the second-best method by 1.94\% in OA and 2.15\% in Kappa. On Salinas, MBTI achieves 96.68\% OA and 96.31\% Kappa, surpassing HyperSIGMA in the two overall metrics while maintaining competitive AA. On FangluTeaFarm, MBTI obtains 95.65\% OA and 93.66\% Kappa, showing clear improvements in overall classification performance. These results indicate that full-band spectral preservation and branch-wise efficient adaptation improve the transfer of hyperspectral foundation models. Classification maps generated by different methods are shown in Figs.~\ref{fig:salinas_classification_maps}--\ref{fig:tf_classification_maps}.

\subsection{Ablation Studies}
The proposed MBTI framework consists of three key components: spectral-continuity-preserving multi-branch preprocessing, spectral multi-branch LoRA, and multi-branch attention-based feature fusion. To verify their contributions, we conduct ablation experiments by removing or replacing individual modules. Four variants are considered:
\begin{enumerate}
\item PCA replacement: the proposed multi-branch spectral preprocessing strategy is replaced by PCA-based dimensionality reduction.
\item Without LoRA: the branch-specific LoRA adapters are removed to evaluate the necessity of parameter-efficient adaptation.
\item Shared LoRA: LoRA parameters with matching layer names are shared across spectral branches, while the branch encoders, spectral grouping, projection layers, fusion module, optimizer, scheduler, and training protocol remain unchanged. This variant is used to evaluate whether branch-specific LoRA parameters are necessary.
\item Without attention: the multi-branch attention fusion module is removed, and features from different branches are directly fused without adaptive recalibration.
\end{enumerate}

\begin{table*}[!t]
\centering
\scriptsize
\caption{Ablation results (\%) on the Salinas, FangluTeaFarm, and KSC datasets.}
\label{tab:ablation_results}
\begin{tabular}{l c c c c c c c c c}
\hline
\multirow{2}{*}{\textbf{Method}}
& \multicolumn{3}{c}{\textbf{Salinas}}
& \multicolumn{3}{c}{\textbf{FangluTeaFarm}}
& \multicolumn{3}{c}{\textbf{KSC}} \\
\cline{2-10}
& \textbf{OA} & \textbf{AA} & \textbf{Kappa}
& \textbf{OA} & \textbf{AA} & \textbf{Kappa}
& \textbf{OA} & \textbf{AA} & \textbf{Kappa} \\
\hline
PCA replacement     & 91.77 & 90.83 & 90.84 & 79.34 & 84.14 & 72.31 & 96.99 & 96.10 & 96.65 \\
Without LoRA        & 91.69 & 92.55 & 90.78 & 94.00 & 93.42 & 91.34 & 99.04 & \textbf{99.29} & 98.93 \\
Shared LoRA   & 92.98 & 94.92 & 92.20 & 88.93 & 93.08 & 84.60 & 93.71 & 94.57 & 93.02 \\
Without attention   & 93.45 & 96.41 & 92.74 & 91.48 & 93.70 & 87.90 & 98.41 & 98.42 & 98.23 \\
MBTI                & \textbf{96.68} & \textbf{97.15} & \textbf{96.31}
                    & \textbf{95.65} & \textbf{94.59} & \textbf{93.66}
                    & \textbf{99.16} & 99.26 & \textbf{99.06} \\
\hline
\end{tabular}
\end{table*}

For fairness, all ablation experiments use the same random seed and experimental settings. The results are reported in Table~\ref{tab:ablation_results}. Overall, removing or replacing any key component leads to performance degradation on most evaluation metrics, which verifies the necessity of the proposed design.

When the proposed multi-branch preprocessing strategy is replaced by PCA, the performance drops clearly on all three datasets. This performance drop may be related to several coupled factors, including the loss of original spectral details, the weakening of local spectral continuity, and the changes in representation basis and input statistical distribution introduced by PCA. In contrast, the proposed grouping and band-reuse strategy preserves the original full-band spectral space and better matches the continuous-band modeling behavior of the hyperspectral foundation model.

Removing the LoRA modules degrades the overall classification performance, especially on Salinas and FangluTeaFarm. This demonstrates that LoRA adaptation is important for learning discriminative task-specific features from different spectral intervals. Furthermore, the shared-LoRA baseline performs worse than the branch-specific LoRA design on all three datasets. For example, on FangluTeaFarm, shared LoRA obtains 88.93\% OA and 84.60\% Kappa, whereas MBTI obtains 95.65\% OA and 93.66\% Kappa. These results indicate that different continuous spectral intervals exhibit distinct characteristics that are better captured and adapted by independent LoRA modules than by a single LoRA module shared across branches. Instead of fully updating the backbone, LoRA provides a parameter-efficient way to adapt the pre-trained model under limited labeled samples.

The complete MBTI framework achieves the best OA and Kappa on all three datasets and obtains the best or highly competitive AA. These results confirm that spectral-continuity-preserving preprocessing, branch-specific LoRA adaptation, and multi-branch attention fusion are complementary. Their combination enables MBTI to achieve a better balance between classification accuracy and efficient downstream adaptation.

\subsection{Analysis of Band-Reuse Strategy}

The proposed multi-branch preprocessing strategy completes the final spectral branch by reusing adjacent bands from the preceding spectral interval when the remaining number of bands is insufficient. To further investigate the role of spectral order, we conduct a controlled band-reuse experiment. In the shuffled-reuse variant, we retain the same reused bands and the same number of input channels as in the proposed method, but randomly permute the reused bands once. Thus, the primary difference between the proposed ordered-reuse strategy and the shuffled-reuse variant is whether the local spectral order is preserved.

\begin{table}[!t]
\centering
\scriptsize
\setlength{\tabcolsep}{3.5pt}
\caption{Effect of band-reuse order (\%).}
\label{tab:band_reuse_order}
\begin{tabular}{l c c c c}
\hline
\textbf{Dataset}
& \textbf{Ordered OA}
& \textbf{Shuffled OA}
& \textbf{Ordered Kappa}
& \textbf{Shuffled Kappa} \\
\hline
KSC
& \textbf{99.16}
& 96.20
& \textbf{99.06}
& 95.77 \\
Salinas
& \textbf{96.68}
& 94.31
& \textbf{96.31}
& 93.68 \\
FangluTeaFarm
& \textbf{95.65}
& 91.32
& \textbf{93.66}
& 87.55 \\
\hline
\end{tabular}
\end{table}

The results in Table~\ref{tab:band_reuse_order} show that randomly permuting the reused bands reduces the OA by 2.96, 2.37, and 4.33 percentage points on KSC, Salinas, and FangluTeaFarm, respectively. The corresponding Kappa values also decrease substantially. These results indicate that the ordered-reuse strategy facilitates the adaptation of HSI data with different numbers of bands to the fixed input structure of the pretrained encoder. The resulting performance improvement is therefore associated with the preservation of local spectral continuity rather than with repeated exposure to a particular spectral interval alone.

\subsection{Analysis of Multi-Branch Feature Fusion Methods}
In MBTI, features from different spectral intervals are extracted independently by multiple branches. Effective integration of these branch-specific representations is therefore critical. To evaluate the influence of fusion strategies, we compare direct concatenation with attention-based fusion mechanisms.

\begin{table*}[!t]
\centering
\scriptsize
\caption{Classification results (\%) of different multi-branch feature fusion methods.}
\label{tab:fusion_methods}
\begin{tabular}{l c c c c c c c c c}
\hline
\multirow{2}{*}{\textbf{Method}}
& \multicolumn{3}{c}{\textbf{Salinas}}
& \multicolumn{3}{c}{\textbf{FangluTeaFarm}}
& \multicolumn{3}{c}{\textbf{KSC}} \\
\cline{2-10}
& \textbf{OA} & \textbf{AA} & \textbf{Kappa}
& \textbf{OA} & \textbf{AA} & \textbf{Kappa}
& \textbf{OA} & \textbf{AA} & \textbf{Kappa} \\
\hline
Concatenation         & 93.45 & 96.41 & 92.74 & 91.48 & 93.70 & 87.90 & 98.41 & 98.42 & 98.23 \\
With self-attention   & 59.24 & 50.50 & 55.62 & 93.47 & \textbf{94.81} & 90.65 & 75.80 & 81.31 & 73.57 \\
With graph attention  & 69.18 & 75.97 & 66.00 & 86.26 & 86.44 & 80.74 & 92.71 & 94.00 & 91.92 \\
MBTI                  & \textbf{96.68} & \textbf{97.15} & \textbf{96.31}
                      & \textbf{95.65} & 94.59 & \textbf{93.66}
                      & \textbf{99.16} & \textbf{99.26} & \textbf{99.06} \\
\hline
\end{tabular}
\end{table*}

The classification results are reported in Table~\ref{tab:fusion_methods}. Direct concatenation treats all branch features equally and ignores the different contributions of spectral intervals. This limits the discriminative ability of the fused representation. Self-attention and graph attention introduce adaptive modeling mechanisms, but they do not consistently improve performance in this setting. The proposed multi-branch attention fusion module achieves the best overall performance, indicating that spectral-branch recalibration is more suitable for integrating the proposed branch features.

\subsection{Parameter Sensitivity Analysis}
This section analyzes two key hyperparameters in MBTI: the number of spectral bands assigned to each branch and the LoRA rank $r$. Since hyperspectral datasets differ in band numbers and scene characteristics, the sensitivity analysis is conducted on KSC, Salinas, and FangluTeaFarm.

\subsubsection{Effect of the Number of Bands per Branch}
The number of bands per branch determines how the original spectrum is divided into continuous spectral groups. A smaller value produces more branches with finer spectral partitioning. A larger value produces fewer branches with wider spectral coverage. As shown in Table~\ref{tab:bands_per_branch}, this trade-off affects classification performance on all three datasets. For KSC and Salinas, assigning 70 bands to each branch achieves the best overall performance, with OA values of 99.16\% and 96.68\%, respectively. For FangluTeaFarm, 35 bands per branch obtains the highest OA and Kappa, while 40 bands per branch achieves the highest AA. These results indicate that excessively fine spectral partitioning does not necessarily improve accuracy. Too many branches increase the difficulty of feature fusion, memory consumption, and optimization. Conversely, overly coarse partitioning may reduce the ability to capture discriminative information from different spectral intervals. Therefore, the final settings are 70 bands per branch for KSC and Salinas, and 35 bands per branch for FangluTeaFarm.

\begin{table}[t]
\centering
\footnotesize
\caption{Classification results (\%) with different numbers of bands per branch.}
\label{tab:bands_per_branch}
\begin{tabular}{l c c c c c}
\hline
\textbf{Dataset} & \textbf{Bands} & \textbf{Branches} & \textbf{OA} & \textbf{AA} & \textbf{Kappa} \\
\hline
\multirow{6}{*}{KSC}
& 30 & 6 & 98.54 & 98.73 & 98.38 \\
& 35 & 6 & 98.46 & 98.69 & 98.29 \\
& 40 & 5 & 99.00 & 99.01 & 98.89 \\
& 50 & 4 & 99.10 & 99.16 & 99.00 \\
& 60 & 3 & 97.95 & 98.30 & 97.72 \\
& 70 & 3 & \textbf{99.16} & \textbf{99.26} & \textbf{99.06} \\
\hline
\multirow{7}{*}{SA}
& 30 & 7 & 96.38 & 96.73 & 95.98 \\
& 35 & 6 & 94.35 & 96.05 & 93.74 \\
& 40 & 6 & 95.77 & 96.66 & 95.30 \\
& 50 & 5 & 96.56 & 96.87 & 96.18 \\
& 60 & 4 & 96.42 & 96.90 & 96.02 \\
& 70 & 3 & \textbf{96.68} & \textbf{97.15} & \textbf{96.31} \\
& 80 & 3 & 95.04 & 95.35 & 94.49 \\
\hline
\multirow{7}{*}{TF}
& 10 & 8 & 94.23 & 94.64 & 91.67 \\
& 15 & 6 & 94.05 & 94.00 & 91.42 \\
& 20 & 4 & 94.36 & 94.52 & 91.85 \\
& 25 & 4 & 93.90 & 94.18 & 91.22 \\
& 30 & 3 & 91.87 & 93.10 & 88.47 \\
& 35 & 3 & \textbf{95.65} & 94.59 & \textbf{93.66} \\
& 40 & 2 & 95.21 & \textbf{95.01} & 93.05 \\
\hline
\end{tabular}
\end{table}

\subsubsection{Effect of the LoRA Rank}
The LoRA rank $r$ controls the dimension of the low-rank adaptation space. A larger rank provides stronger adaptation capacity, but it also introduces more trainable parameters. To evaluate its influence, $r$ is selected from $\{2,4,8,16,32,64,128\}$, while the number of bands per branch is fixed to the optimal setting of each dataset. As shown in Table~\ref{tab:lora_rank}, the influence of $r$ is dataset-dependent, and the improvement is not monotonic. KSC and FangluTeaFarm obtain the best OA and Kappa at $r=128$, whereas Salinas achieves the best OA and Kappa at $r=8$. Larger ranks enlarge the task-specific adaptation capacity, but they also substantially increase the parameter cost. For example, on KSC, the trainable parameters increase from 3.76M at $r=2$ to 61.48M at $r=128$, while the OA gain is limited. Considering both accuracy and efficiency, $r=8$ is adopted as the default setting because it achieves competitive performance with substantially fewer trainable parameters.

\begin{table}[t]
\centering
\footnotesize
\caption{Classification results (\%) with different LoRA ranks.}
\label{tab:lora_rank}
\begin{tabular}{l c c c c c}
\hline
\textbf{Dataset} & \textbf{Rank} & \textbf{OA} & \textbf{AA} & \textbf{Kappa} & \textbf{Trainable Params} \\
\hline
\multirow{7}{*}{KSC}
& 2   & 98.93 & 98.89 & 98.80 & 3.76 M \\
& 4   & 98.48 & 98.55 & 98.31 & 4.67 M \\
& 8   & 99.16 & 99.26 & 99.06 & 6.51 M \\
& 16  & 98.29 & 97.84 & 98.10 & 10.17 M \\
& 32  & 99.16 & 99.14 & 99.06 & 17.50 M \\
& 64  & 98.31 & 97.78 & 98.12 & 32.16 M \\
& 128 & \textbf{99.29} & \textbf{99.37} & \textbf{99.21} & 61.48 M \\
\hline
\multirow{7}{*}{SA}
& 2   & 95.77 & 96.55 & 95.30 & 3.76 M \\
& 4   & 96.06 & 96.87 & 95.62 & 4.68 M \\
& 8   & \textbf{96.68} & 97.15 & \textbf{96.31} & 6.51 M \\
& 16  & 96.06 & 96.62 & 95.62 & 10.18 M \\
& 32  & 96.57 & 97.05 & 96.19 & 17.51 M \\
& 64  & 96.21 & \textbf{97.21} & 95.79 & 32.17 M \\
& 128 & 96.45 & 96.92 & 96.05 & 61.49 M \\
\hline
\multirow{7}{*}{TF}
& 2   & 95.67 & \textbf{95.19} & 93.70 & 3.75 M \\
& 4   & 95.01 & 94.88 & 92.77 & 4.65 M \\
& 8   & 95.65 & 94.59 & 93.66 & 6.46 M \\
& 16  & 95.73 & 94.97 & 93.78 & 10.07 M \\
& 32  & 95.54 & 94.90 & 93.51 & 17.29 M \\
& 64  & 95.73 & 94.87 & 93.77 & 31.74 M \\
& 128 & \textbf{95.86} & 94.88 & \textbf{93.96} & 60.62 M \\
\hline
\end{tabular}
\end{table}

\subsection{Efficiency and Cross-Framework Inference Analysis}

To further analyze the practical efficiency and deployment potential of MBTI, we conduct two additional experiments on the FangluTeaFarm dataset. The first experiment compares MBTI with representative hyperspectral foundation-model-based methods in terms of classification accuracy, trainable parameters, training-state memory, and inference time. The second experiment evaluates inference behavior of MBTI under different deep learning frameworks.

\begin{table}[!t]
\centering
\footnotesize
\caption{Efficiency comparison of foundation-model-based methods on the FangluTeaFarm dataset.}
\label{tab:foundation_efficiency}
\begin{tabular}{l c c c c}
\hline
\textbf{Method} & \textbf{OA} & \textbf{Trainable Params} & \textbf{Train State} & \textbf{Infer. Time} \\
 & \textbf{(\%)} & \textbf{(M)} & \textbf{(MiB)} & \textbf{(s)} \\
\hline
HyperSIGMA & 88.53 & 194.24 & 2963.85 & 0.49 \\
HyperFree  & 92.86 & 0.00   & 0.00    & 38.12 \\
MBTI       & \textbf{95.65} & 6.46 & 98.53 & 1.13 \\
\hline
\end{tabular}
\end{table}

\subsubsection{Foundation-model efficiency}
As shown in Table~\ref{tab:foundation_efficiency}, MBTI achieves the highest OA among the compared foundation-model-based methods while requiring far fewer trainable parameters than full fine-tuning. The branch encoders are initialized from the same HyperSIGMA checkpoint, and their pretrained base weights are frozen during downstream adaptation. Only the branch-specific LoRA modules, branch projection layers, fusion module, and classifier are optimized. Therefore, the efficiency advantage of MBTI mainly lies in trainable-parameter efficiency and optimization-state memory reduction. We also note that MBTI is not designed to minimize raw inference latency. Since MBTI executes the encoder for multiple spectral branches, its inference time is longer than that of single-branch HyperSIGMA. This is an accuracy-efficiency trade-off: MBTI improves classification accuracy and greatly reduces downstream optimization cost, at the expense of a moderate increase in inference time.

Compared with HyperFree, MBTI achieves higher classification accuracy and much faster inference. Although HyperFree avoids conventional fine-tuning and therefore does not introduce trainable parameters during downstream adaptation, its tuning-free adaptation strategy leads to a relatively heavy inference process. In contrast, MBTI performs task-specific adaptation through lightweight LoRA modules during training and then uses a standard forward inference pipeline. Therefore, compared with HyperFree, MBTI provides a better balance between classification accuracy, adaptation cost, and inference latency.

\subsubsection{Cross-framework inference}

To examine the deployment characteristics of MBTI across different deep learning frameworks, we further evaluate it under Jittor in addition to PyTorch. Specifically, the final PyTorch checkpoint is converted into a Jittor-readable format and tested with a Jittor-based inference implementation on the FangluTeaFarm dataset. This experiment is designed to assess whether the trained MBTI checkpoint can preserve its accuracy while achieving lower memory consumption after framework conversion.

\begin{figure}[!t]
\centering
\includegraphics[width=\columnwidth]{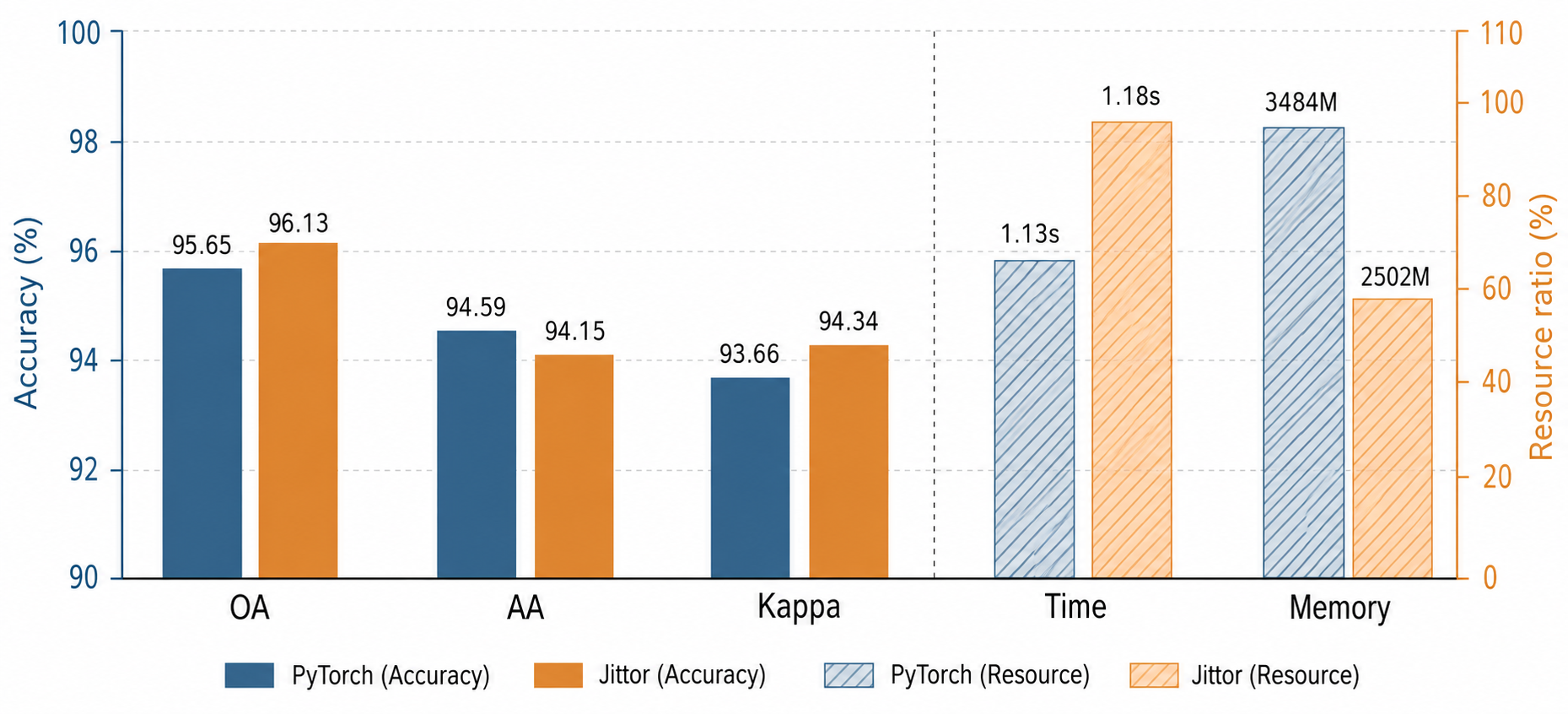}
\caption{Cross-framework inference comparison between PyTorch and Jittor on the FangluTeaFarm dataset.}
\label{fig:cross_framework_inference}
\end{figure}

The cross-framework inference results are shown in Fig.~\ref{fig:cross_framework_inference}. The Jittor implementation achieves comparable classification performance to the PyTorch implementation. Specifically, Jittor obtains slightly higher OA and Kappa, while its AA is slightly lower. These small differences may be caused by framework-level numerical differences, operator implementation details, and the conversion of model weights between PyTorch and Jittor. Overall, the results indicate that the converted Jittor inference pipeline can preserve the main discriminative capability of the trained MBTI model.

In terms of resource consumption, Jittor shows a clear advantage in peak GPU memory usage. On FangluTeaFarm, the peak GPU memory decreases from 3484 MiB in PyTorch to 2502 MiB in Jittor. This suggests that Jittor has potential for memory-efficient deployment of MBTI, especially when full-scene inference needs to be performed on devices with limited GPU memory. Meanwhile, the inference time of Jittor is slightly longer than that of PyTorch, increasing from 1.13 s to 1.18 s. This may be related to differences in graph compilation, runtime scheduling, and operator-level optimization between the two frameworks. In addition, although the model parameters are converted into a Jittor-compatible format, some modules originally developed and optimized in PyTorch may not achieve identical execution efficiency after being reimplemented with Jittor operators.

Overall, these results demonstrate that MBTI provides parameter-efficient downstream adaptation while retaining practical cross-framework portability. Compared with full fine-tuning, MBTI substantially reduces the number of trainable parameters and the associated optimization-state memory. After conversion to Jittor, it maintains classification performance comparable to the PyTorch implementation and further reduces peak GPU memory.

\section{Conclusion}
This paper proposed MBTI, a multi-branch efficient fine-tuning framework for hyperspectral image classification based on hyperspectral foundation models. Instead of compressing or selecting bands to satisfy fixed model inputs, MBTI preserves the full-band spectrum by dividing HSIs into continuous spectral branches and reusing adjacent bands when the final branch is incomplete. This design reduces spectral information loss and maintains local spectral continuity during downstream adaptation. By inserting branch-specific LoRA modules into the frozen pretrained backbone, MBTI adapts different spectral intervals with only a small fraction of trainable parameters, while the spectral-channel recalibration module further integrates branch features according to their discriminative contributions. Experiments on KSC, Salinas, and FangluTeaFarm show that MBTI consistently outperforms a wide range of representative methods. Ablation studies further confirm the effectiveness of each component in improving adaptation efficiency and deployment flexibility. Future work will investigate more general spectral alignment and cross-sensor knowledge transfer, together with native Jittor training and framework-specific operator optimization, to improve adaptation across heterogeneous sensors and enable more efficient cross-framework deployment.


\begin{thebibliography}{40}


\bibitem{ref1}
N. Audebert, B. Le Saux, and S. Lef\`evre, ``Deep learning for classification of hyperspectral data: A comparative review,'' \textit{IEEE Geoscience and Remote Sensing Magazine}, vol. 7, no. 2, pp. 159--173, 2019.

\bibitem{ref2}
S. Li, W. Song, L. Fang, Y. Chen, P. Ghamisi, and J. A. Benediktsson, ``Deep learning for hyperspectral image classification: An overview,'' \textit{IEEE Transactions on Geoscience and Remote Sensing}, vol. 57, no. 9, pp. 6690--6709, 2019.

\bibitem{ref3}
Q. Zhu, M. Xu, R. Ma, L. Ran, J. Xue, and Q. Guan, ``Knowledge-data-model-driven multimodal few-shot learning for hyperspectral fine classification: Generalization across sensor, category and scene,'' \textit{ISPRS Journal of Photogrammetry and Remote Sensing}, vol. 233, pp. 623--650, 2026.

\bibitem{ref4}
G. Mercier and M. Lennon, ``Support vector machines for hyperspectral image classification with spectral-based kernels,'' in \textit{Proc. IGARSS 2003 IEEE International Geoscience and Remote Sensing Symposium}, 2003, pp. 288--290.

\bibitem{ref5}
Y. Zhang, G. Cao, X. Li, and B. Wang, ``Cascaded random forest for hyperspectral image classification,'' \textit{IEEE Journal of Selected Topics in Applied Earth Observations and Remote Sensing}, vol. 11, no. 4, pp. 1082--1094, 2018.

\bibitem{ref6}
F. Zhou, R. Hang, Q. Liu, and X. Yuan, ``Hyperspectral image classification using spectral-spatial LSTMs,'' \textit{Neurocomputing}, vol. 328, pp. 39--47, 2019.

\bibitem{ref7}
Z. Zhong, J. Li, Z. Luo, and M. Chapman, ``Spectral--spatial residual network for hyperspectral image classification: A 3-D deep learning framework,'' \textit{IEEE Transactions on Geoscience and Remote Sensing}, vol. 56, no. 2, pp. 847--858, 2017.

\bibitem{ref8}
D. Hong, Z. Han, J. Yao, L. Gao, B. Zhang, A. Plaza, and J. Chanussot, ``SpectralFormer: Rethinking hyperspectral image classification with transformers,'' \textit{IEEE Transactions on Geoscience and Remote Sensing}, vol. 60, pp. 1--15, 2021.

\bibitem{ref9}
B. Pan, Z. Shi, and X. Xu, ``MugNet: Deep learning for hyperspectral image classification using limited samples,'' \textit{ISPRS Journal of Photogrammetry and Remote Sensing}, vol. 145, pp. 108--119, 2018.

\bibitem{ref10}
Q. Zhu, H. Li, W. Deng, Q. Guan, and J. Luo, ``From intra-distinctiveness to inter-invariance: A cycle-resemblance few-shot transformation network for cross-domain hyperspectral image classification,'' \textit{IEEE Transactions on Geoscience and Remote Sensing}, vol. 63, pp. 1--16, 2024.

\bibitem{ref11}
Y. Wang, Q. Yao, J. T. Kwok, and L. M. Ni, ``Generalizing from a few examples: A survey on few-shot learning,'' \textit{ACM Computing Surveys}, vol. 53, no. 3, pp. 1--34, 2020.

\bibitem{ref12}
B. Liu, X. Yu, A. Yu, P. Zhang, G. Wan, and R. Wang, ``Deep few-shot learning for hyperspectral image classification,'' \textit{IEEE Transactions on Geoscience and Remote Sensing}, vol. 57, no. 4, pp. 2290--2304, 2018.

\bibitem{ref13}
B. Wang, Y. Xu, Z. Wu, T. Zhan, and Z. Wei, ``Spatial--spectral local domain adaption for cross domain few shot hyperspectral images classification,'' \textit{IEEE Transactions on Geoscience and Remote Sensing}, vol. 60, pp. 1--15, 2022.

\bibitem{ref14}
A. Qin, C. Yuan, Q. Li, X. Luo, F. Yang, T. Song, and C. Gao, ``Few-shot learning with prototype rectification for cross-domain hyperspectral image classification,'' \textit{IEEE Transactions on Geoscience and Remote Sensing}, vol. 62, pp. 1--15, 2024.

\bibitem{ref15}
Q. Liu, J. Peng, N. Chen, W. Sun, Y. Ning, and Q. Du, ``Category-specific prototype self-refinement contrastive learning for few-shot hyperspectral image classification,'' \textit{IEEE Transactions on Geoscience and Remote Sensing}, vol. 61, pp. 1--16, 2023.

\bibitem{ref16}
J. Sun, C. Zheng, E. Xie, Z. Liu, R. Chu, J. Qiu, J. Xu, M. Ding, H. Li, and M. Geng, ``A survey of reasoning with foundation models: Concepts, methodologies, and outlook,'' \textit{ACM Computing Surveys}, vol. 57, no. 11, pp. 1--43, 2025.

\bibitem{ref17}
N. Ding, Y. Qin, G. Yang, F. Wei, Z. Yang, Y. Su, S. Hu, Y. Chen, C.-M. Chan, and W. Chen, ``Parameter-efficient fine-tuning of large-scale pre-trained language models,'' \textit{Nature Machine Intelligence}, vol. 5, no. 3, pp. 220--235, 2023.

\bibitem{ref18}
D. Wang, M. Hu, Y. Jin, Y. Miao, J. Yang, Y. Xu, X. Qin, J. Ma, L. Sun, and C. Li, ``HyperSIGMA: Hyperspectral intelligence comprehension foundation model,'' \textit{IEEE Transactions on Pattern Analysis and Machine Intelligence}, 2025.

\bibitem{ref19}
W. Kong, B. Liu, X. Bi, C. Yu, X. Li, and Y. Chen, ``HyperSL: A spectral foundation model for hyperspectral image interpretation,'' \textit{IEEE Transactions on Geoscience and Remote Sensing}, vol. 63, Art. no. 5513119, 2025, doi: 10.1109/TGRS.2025.3566205.

\bibitem{ref20}
X. Zhao, Z. Xiong, and X. X. Zhu, ``HySens: Sensor-agnostic foundation models for hyperspectral data,'' \textit{IEEE Transactions on Geoscience and Remote Sensing}, vol. 64, Art. no. 5513915, 2026, doi: 10.1109/TGRS.2026.3691782.

\bibitem{ref21}
J. Li, Y. Liu, X. Wang, Y. Peng, C. Sun, S. Wang, Z. Sun, T. Ke, X. Jiang, T. Lu, A. Zhao, and Y. Zhong, ``HyperFree: A channel-adaptive and tuning-free foundation model for hyperspectral remote sensing imagery,'' in \textit{Proc. IEEE/CVF Conference on Computer Vision and Pattern Recognition}, 2025, pp. 23048--23058.

\bibitem{ref22}
S. Lu, J. Guo, J. R. Zimmer-Dauphinee, J. M. Nieusma, X. Wang, S. A. Wernke, and Y. Huo, ``Vision foundation models in remote sensing: A survey,'' \textit{IEEE Geoscience and Remote Sensing Magazine}, 2025.

\bibitem{ref23}
N. A. A. Braham, C. M. Albrecht, J. Mairal, J. Chanussot, Y. Wang, and X. X. Zhu, ``SpectralEarth: Training hyperspectral foundation models at scale,'' \textit{IEEE Journal of Selected Topics in Applied Earth Observations and Remote Sensing}, vol. 18, pp. 16780--16797, 2025, doi: 10.1109/JSTARS.2025.3581451.

\bibitem{ref24}
Z. H. Tushar and S. Purushotham, ``HyperFM: An efficient hyperspectral foundation model with spectral grouping,'' in \textit{Proc. IEEE/CVF Conference on Computer Vision and Pattern Recognition Workshops (CVPRW)}, 2026, pp. 6955--6964.

\bibitem{ref25}
Z. Han, C. Gao, J. Liu, J. Zhang, and S. Q. Zhang, ``Parameter-efficient fine-tuning for large models: A comprehensive survey,'' \textit{arXiv preprint arXiv:2403.14608}, 2024.

\bibitem{ref26}
Y. Xin, J. Yang, S. Luo, Y. Du, Q. Qin, K. Cen, Y. He, Z. Zhang, B. Fu, and X. Yang, ``Parameter-efficient fine-tuning for pre-trained vision models: A survey and benchmark,'' \textit{arXiv preprint arXiv:2402.02242}, 2024.

\bibitem{ref27}
L. Xu, H. Xie, S. J. Qin, X. Tao, and F. L. Wang, ``Parameter-efficient fine-tuning methods for pretrained language models: A critical review and assessment,'' \textit{IEEE Transactions on Pattern Analysis and Machine Intelligence}, 2026.

\bibitem{ref28}
L. Hu, H. Yu, W. Lu, D. Yin, X. Sun, and K. Fu, ``AiRs: Adapter in remote sensing for parameter-efficient transfer learning,'' \textit{IEEE Transactions on Geoscience and Remote Sensing}, vol. 62, Art. no. 5605218, 2024, doi: 10.1109/TGRS.2024.3351889.

\bibitem{ref29}
Z. Dong, Y. Gu, and T. Liu, ``UPetu: A unified parameter-efficient fine-tuning framework for remote sensing foundation model,'' \textit{IEEE Transactions on Geoscience and Remote Sensing}, vol. 62, Art. no. 5616613, 2024, doi: 10.1109/TGRS.2024.3382734.

\bibitem{ref30}
Y. Zhang, W. Li, M. Zhang, J. Han, R. Tao, and S. Liang, ``SpectralX: Parameter-efficient domain generalization for spectral remote sensing foundation models,'' arXiv:2508.01731, 2025.

\bibitem{ref31}
B. Ligan, K. Jbilou, F. Kalloubi, and A. Ratnani, ``Parameter-efficient fine-tuning of multispectral foundation models for hyperspectral image classification,'' arXiv:2505.15334, 2025.

\bibitem{ref32}
M. A. Hearst, S. T. Dumais, E. Osuna, J. Platt, and B. Sch\"olkopf, ``Support vector machines,'' \textit{IEEE Intelligent Systems and Their Applications}, vol. 13, no. 4, pp. 18--28, 1998.

\bibitem{ref33}
H. Lee and H. Kwon, ``Going deeper with contextual CNN for hyperspectral image classification,'' \textit{IEEE Transactions on Image Processing}, vol. 26, no. 10, pp. 4843--4855, 2017.

\bibitem{ref34}
R. Li, S. Zheng, C. Duan, Y. Yang, and X. Wang, ``Classification of hyperspectral image based on double-branch dual-attention mechanism network,'' \textit{Remote Sensing}, vol. 12, no. 3, p. 582, 2020.

\bibitem{ref35}
W. Ma, Q. Yang, Y. Wu, W. Zhao, and X. Zhang, ``Double-branch multi-attention mechanism network for hyperspectral image classification,'' \textit{Remote Sensing}, vol. 11, no. 11, p. 1307, 2019.

\bibitem{ref36}
W. Wang, S. Dou, Z. Jiang, and L. Sun, ``A fast dense spectral--spatial convolution network framework for hyperspectral images classification,'' \textit{Remote Sensing}, vol. 10, no. 7, p. 1068, 2018.

\bibitem{ref37}
Z. Li, M. Liu, Y. Chen, Y. Xu, W. Li, and Q. Du, ``Deep cross-domain few-shot learning for hyperspectral image classification,'' \textit{IEEE Transactions on Geoscience and Remote Sensing}, vol. 60, pp. 1--18, 2022.

\bibitem{ref38}
Y. Zhang, W. Li, M. Zhang, S. Wang, R. Tao, and Q. Du, ``Graph information aggregation cross-domain few-shot learning for hyperspectral image classification,'' \textit{IEEE Transactions on Neural Networks and Learning Systems}, vol. 35, no. 2, pp. 1912--1925, 2022.

\bibitem{ref39}
Y. Wang, M. Liu, Y. Yang, Z. Li, Q. Du, Y. Chen, F. Li, and H. Yang, ``Heterogeneous few-shot learning for hyperspectral image classification,'' \textit{IEEE Geoscience and Remote Sensing Letters}, vol. 19, pp. 1--5, 2021.

\bibitem{ref40}
J. Hu, L. Shen, and G. Sun, ``Squeeze-and-excitation networks,'' in \textit{Proc. IEEE/CVF Conf. Comput. Vis. Pattern Recognit. (CVPR)}, 2018, pp. 7132--7141.

\end{thebibliography}
\end{document}